\documentclass[10pt,journal,compsoc]{IEEEtran}
\ifCLASSOPTIONcompsoc
  \usepackage[nocompress]{cite}
\else
  \usepackage{cite}
\fi
\ifCLASSINFOpdf
\else
\fi
\usepackage{amsmath,amsfonts}
\usepackage{algorithmic}
\usepackage{array}
\usepackage{textcomp}
\usepackage{stfloats}
\usepackage{url}
\usepackage{verbatim}
\usepackage{graphicx}
\usepackage{amssymb}
\usepackage{bm}
\usepackage{subfigure}
\usepackage{multirow}
\usepackage{color}
\usepackage{cite}
\usepackage{tabularx}
\usepackage{makecell}
\usepackage{hyperref}

\begin{document}

\title{TDMD: A Database for Dynamic Color Mesh Subjective and Objective Quality Explorations}

%
%
%
%

\author{ Qi Yang,
        Joel Jung, 
        Timon Deschamps,
        Xiaozhong Xu,~\IEEEmembership{Member,~IEEE},
        and Shan Liu, ~\IEEEmembership{Fellow,~IEEE}

\thanks{Q. Yang, J. Jung, T. Deschamps, X. Xu, and S. Liu are from Tencent Media Lab, (e-mail: \{chinoyang, joeljung, timond, xiaozhongxu, shanl\}@tencent.com) }

}

\IEEEtitleabstractindextext{%
\begin{abstract}

Dynamic colored meshes (DCM) are widely used in various applications; however, these meshes may undergo different processes, such as compression or transmission, which can distort them and degrade their quality. To facilitate the development of objective metrics for DCMs and study the influence of typical distortions on their perception, we create the Tencent - dynamic colored mesh database (TDMD) containing eight reference DCM objects with six typical distortions. Using processed video sequences (PVS) derived from the DCM, we have conducted a large-scale subjective experiment that resulted in 303 distorted DCM samples with mean opinion scores, making the TDMD the largest available DCM database to our knowledge. This database enabled us to study the impact of different types of distortion on human perception and offer recommendations for DCM compression and related tasks. Additionally, we have evaluated three types of state-of-the-art objective metrics on the TDMD, including image-based, point-based, and video-based metrics, on the TDMD. Our experimental results highlight the strengths and weaknesses of each metric, and we provide suggestions about the selection of metrics in practical DCM applications. The TDMD will be made publicly available at the following location: {\url{https://multimedia.tencent.com/resources/tdmd}}.

\end{abstract}

}

\maketitle

\IEEEdisplaynontitleabstractindextext

%
\IEEEpeerreviewmaketitle


\section{Introduction}\label{sec:intro}

As a typical representation of three-dimensional (3D) graphics, 3D mesh plays a crucial role in various applications \cite{attene2013polygon}, such as animated advertisement, digital entertainment, and education. These applications generally require high quality meshes to provide better quality of experience through improved immersion. Thus, accurately evaluating the influence of distortions on mesh quality is an essential task.  

Most of the current work on mesh quality assessment focuses on static meshes~\cite{abouelaziz2017mesh}, and sometimes disregards the color information. However, dynamic colored meshes (DCMs) have become increasingly popular, and dedicated compression algorithms have attracted considerable attention in the WG 7 - MPEG Coding of 3D Graphics and Haptics ~\cite{MPEG-MESH-cfp}. Conducting comprehensive DCM quality assessment research is important to control tasks such as lossy compression ~\cite{meshcomp-maglo20153d}, transmission ~\cite{mesht-yang2004progressive}, reconstruction~\cite{meshre-jiang20183d}, and enhancement ~\cite{meshen-hansen2005mesh}.  

DCM can be classified into two different categories depending on how color information is represented. In the first category, the color information is stored in a texture map, and a group of UV coordinates is used to indicate the texture of the mesh samples. In the second category, the color information is stored per vertex in a similar way to colored point cloud.  Since the first type is more commonly used, our focus in this paper is on studying DCM using texture maps.

Studies on quality assessment of DCMs typically involve two types of evaluations: subjective and objective. To elaborate on the rationale for our work, we begin by discussing the issues associated with each of them. 

The main purpose of DCM subjective assessment is to study the influence of mesh distortions on human perception. Regarding the viewing environment, the DCM can either be rendered and presented on a two-dimensional screen, or it can be rendered via a virtual reality (VR) headset and presented in an immersive space. Currently, most subjective assessment studies focus on static meshes~\cite{meshcompression-christaki2019subjective, DWPM-corsini2007watermarked, guo2016subjective}. 
To contribute to the development of objective metrics, they investigate how humans perceive mesh degradations in different viewing environments. However, there is a lack of evidence on whether the conclusions obtained on static meshes remain valid on DCM. Although ~\cite{dy-torkhani2015perceptual} investigates dynamic meshes with a 2D monitor viewing environment, the proposed samples are non-colored meshes, and color information tends to have a masking effect for artifacts on subjective perception ~\cite{TMM-masking}. The lack of studies on the responses of human perception to different DCM distortions is consequently the first motivation for providing in this paper a new DCM database. 

The main purpose of DCM objective assessment is to propose effective metrics that have a high correlation with human perception and can replace the expensive and time-consuming subjective assessment in downstream DCM processing and applications. Currently, there are few metrics specifically designed for DCM, since most mesh-oriented metrics are designed for uncolored static mesh (e.g., ~\cite{MSDM-lavoue2006perceptually,DWPM-corsini2007watermarked,dame-vavsa2012dihedral}). Furthermore, these metrics require the mesh samples to share the same connectivity, same vertex density, or the same level of details, otherwise, they cannot extract features and output objective scores. These requirements limit the utilization of these metrics, especially since the possible distortions of mesh samples can be complex and diverse. For example, compression can lead to the overlap of mesh vertices, and reconstruction can generate mesh samples with holes or cracks. The DCM samples used in WG 7 ~\cite{MPEG-MESH-Coding}, which are representations of human models, do not meet the requirements of the above metrics.  This indicates that above listed metrics cannot be used to evaluate the quality of DCM which are under study and application in standardization organizations. \textcolor{black}{Although WG 7 ~\cite{MPEG-MESH-metric} proposed two strategies, image-based and point-based metrics, as substitute to measure DCM quality, these metrics are not initially designed for meshes. The lack of reliable and applicable objective quality metrics for DCM is another motivation of this paper for providing in this paper a new DCM database.}

To study the influence of mesh processing algorithms on DCM perception and validate whether objective evaluation strategies are effective for DCMs, a major contribution of this paper is to create and release a new DCM database, called Tencent - DCM database (TDMD). TDMD contains eight reference meshes and six types of distortions, namely color noise (CN), texture map downsampling (DS), geometrical Gaussian noise (GN), mesh decimation (MD), MPEG lossy compression (MLC), and texture map compression (TC). Each distortion is applied with five to seven degrees of severity. We convert the DCM samples into processed video sequences (PVS) with a predefined camera path and conduct large-scale subjective experiments to obtain mean opinion scores (MOS).  TDMD contains MOS for 303 distorted DCM samples which makes it, as far as we know,  the largest publicly available DCM database. On the basis of this proposed database, the second contribution of this paper is to analyze the impact of different distortions on human perception. The third contribution consists in testing three types of objective metrics on the proposed database. Besides the two kinds proposed by the WG 7, we additionally use PVSs to test extra image/video quality assessment metrics, such as SSIM \cite{wang2004image} and VMAF \cite{vmaf-li2016toward}.  We label this category as the video-based metric. Two correlation indicators are calculated to reveal the performance of the objective metrics: the Pearson linear correlation coefficient (PLCC) and the Spearman rank-order correlation coefficient (SROCC). 
Moreover, point-based metrics rely on mesh sampling and little research has been conducted on the influence of sampling.  Therefore, the last contribution of this paper is to study the influence of sampling on the performance of point-based metrics.
Based on the experiment results, we analyze the advantages and disadvantages of each metric and provide suggestions for metric application. 

The remainder of this paper is laid out as follows. Section \ref{sec:related work} presents the related work about mesh quality assessment. Section \ref{sec:sub-data-con} details the construction of the TDMD database and analyzes the characteristics of distortion in human perception. Section \ref{sec:obj-me-re} studies the performance of different types of objective metrics on the proposed database. Finally, section \ref{sec:conclusion} concludes the paper and highlights future work.
\section{Related Work}\label{sec:related work}

In this section, we summarize the state of the art of mesh quality assessment.
\subsection{Subjective Mesh Quality Assessment}
Mesh subjective quality assessment has been studied for many years. Initially, researchers focused on the artifacts of non-colored meshes, including compression, watermarking, surface noise, and simplification. Specifically, ~\cite{meshcompression-christaki2019subjective} collected four 3D human reconstructions and four scanned objects as references, and selected three different codecs to generate distorted samples. A VR application was used to realize pairwise testings to evaluate the visual quality of compressed meshes. In ~\cite{DWPM-corsini2007watermarked}, a 2D monitor was used to present four mesh objects with watermarking to a panel of eleven subjects, who were asked to score the samples. ~\cite{dy-torkhani2015perceptual} also used a 2D monitor to display meshes with compression and surface noise distortions. Ten non-colored dynamic meshes were used, and the authors proposed simulated distortion (e.g., surface noise) and real-world distortion (e.g., compression, network transmission error).

With the increasing application of meshes in various fields, such as digital entertainment and computer games, there have been an increasing number of studies focusing on colored meshes. ~\cite{guo2016subjective} introduced a database containing five colored meshes with texture maps as references. Five types of distortions were proposed and a video-based subjective methodology was adopted to collect scores. ~\cite{nehme2020visual} used a VR environment to score five mesh samples with four types of distortions. The color information of these meshes was stored per vertex (as opposed to using a texture map). In ~\cite{nehme2022textured}, subjective experiments were conducted  using crowdsourcing and four types of distortions were used for meshes with texture maps.  

We list published mesh databases in Table \ref{tab:mesh-database}. 
\textcolor{black}{In summary, \cite{meshcompression-christaki2019subjective,DWPM-corsini2007watermarked,lavoue2006perceptually,guo2016subjective,nehme2020visual,nehme2022textured} proposed static mesh databases, \cite{dy-torkhani2015perceptual} created a dynamic non-colored mesh database. Therefore}, no DCM database is currently available, and consequently,  the influence of DCM typical distortions on perceived quality is unclear. Therefore, there is a need to construct a new database, involving state-of-the-art (SOTA) mesh distortions (e.g., mesh decimation, lossy compression algorithms) to facilitate the study of DCM subjective quality assessment.   

     \begin{table}[!ht]   
	\centering
	\caption{Mesh database survey.} \label{tab:mesh-database}
     \begin{scriptsize}
 \renewcommand{\arraystretch}{1.5}
	\setlength{\tabcolsep}{0.8mm}{
	\begin{tabular}{|c|c|c|c|c|}
		\hline
		Name & Type & Color & Scale & Distortions \\ \cline{1-5}
  MMM2019\cite{meshcompression-christaki2019subjective}&Static&No&88&2\\ \cline{1-5}
  TMM2006\cite{DWPM-corsini2007watermarked}&Static&No&44&3\\ \cline{1-5}
  LIRIS GPD\cite{lavoue2006perceptually} & Static & No & 88 & 2\\ \cline{1-5}
  LIRIS TDMD\cite{guo2016subjective}&Static&Yes&136&5\\ \cline{1-5}
  TVCG2021\cite{nehme2020visual}&Static&Yes&80&4\\ \cline{1-5}
  TOG2023\cite{nehme2022textured}&Static&Yes&\thead{343K \\ (Pseudo MOS)}&5\\ \cline{1-5}
  SPIC2015\cite{dy-torkhani2015perceptual}&Dynamic&No&276&4\\ \cline{1-5}
  TDMD&Dynamic&Yes&303&6 \\ \hline
		
	\end{tabular}}
 \end{scriptsize}
\end{table}

\subsection{Objective Mesh Quality Assessment}
In a similar fashion to the development of mesh subjective quality assessment, early mesh objective metrics were developed for non-colored meshes. The simplest metrics are distance-based, such as the root mean square error (RMSE) ~\cite{rms-cignoni1998metro} and the Hausdorff distance (HD) ~\cite{hd-aspert2002mesh}.  RMSE calculates the distances between the vertices in the reference mesh and the corresponding vertices in the distorted mesh, and then averages the distances as the objective score. HD first measures the distance between the points from the reference surface and the corresponding points from the distorted surface, and then selects the maximum value as the objective score. Similarly to the mean square error (MSE) and peak signal-to-noise ratio (PSNR)  in the case of image quality assessment (IQA), these distance-based metrics have a low correlation with human perception and generally fail to reflect the influence of distortion on perceived quality. 

Inspired by the achievements of SSIM ~\cite{wang2004image} in the IQA field, researchers proposed using structural features to quantify mesh quality. ~\cite{MSDM-lavoue2006perceptually} suggested leveraging surface curvature to infer mesh geometrical distortion. ~\cite{DWPM-corsini2007watermarked} introduced global roughness to measure mesh surface distortion. ~\cite{GL-karni2000spectral} made use of Geometric Laplacian to more accurately capture subtle visual distortions resulting from lossy compression. ~\cite{dame-vavsa2012dihedral} integrated dihedral angle and a masking weight to predict mesh quality. These metrics have a strong relationship with the topological characteristics of the mesh and show better performance than distance-based metrics ~\cite{abouelaziz2017mesh}. However, they have strict constraints regarding the meshes' geometrical topology. For meshes having duplicated vertices, surface holes, or isolated faces, these metrics cannot extract features and fail to predict objective quality. 

To solve the above problem, two strategies have been proposed. The first converts meshes into images through projection, then uses IQA metrics to predict their quality. Specifically, ~\cite{pro-lavoue2015efficiency}  conducted a comprehensive study involving seven 2D metrics (among which MSE, SSIM, and FSIM ~\cite{fsim-zhang2011fsim}) and illustrated their benefits for various applications. ~\cite{MPEG-MESH-metric} proposed to use Fibonacci sphere lattice ~\cite{fibonacci} to generate 16 viewpoints for projection. They then used depth and color information to predict quality. Besides, with the development of machine learning and deep neural network, many no-reference image-based mesh quality metrics (i.e.,  metrics that  only rely on the assessed distorted sample to be computed) have been proposed, such as ~\cite{p-nr2-abouelaziz20203d,p-n4-abouelaziz2020combination,p-n5-abouelaziz2018convolutional,nehme2022textured}. An advantage of this strategy is that it converts the challenging mesh quality evaluation problem into a well-researched image quality assessment problem. Consequently, many effective IQA metrics can be employed to predict the objective score. This strategy also presents obvious drawbacks, such as the necessity for rendering before capturing images and the influence of viewpoint selection on final results ~\cite{yang2020predicting}.

The second approach is to convert meshes into point clouds (via mesh surface sampling), then use point cloud objective metrics to assess the quality. Several sampling methods have been proposed, such as grid sampling, face sampling, and surface subdivision sampling ~\cite{MPEG-MESH-metric}. Point cloud-based methods allow quality assessment for all kinds of meshes without restrictions and a number of effective colored point cloud metrics \cite{meynet2020pcqm,yang2020inferring,Yang_2022_CVPR,yang2021mped,shanTVCG,suTIP,liuTIP,liuTVCG} that can be chosen from. However, different sampling methods might generate different point clouds for the same mesh, which influences the results of the quality assessment. 

Therefore, there is a need to construct a new database, involving state-of-the-art mesh distortions to facilitate the study of DCM objective quality metrics.

\section{Database Construction}\label{sec:sub-data-con}

In this part, we introduce the construction of the TDMD, including content description, distortion generation, video generation, and subjective testing methodology. 

\subsection{Content Description}
\begin{figure}
    \centering
    \includegraphics[width=1\linewidth]{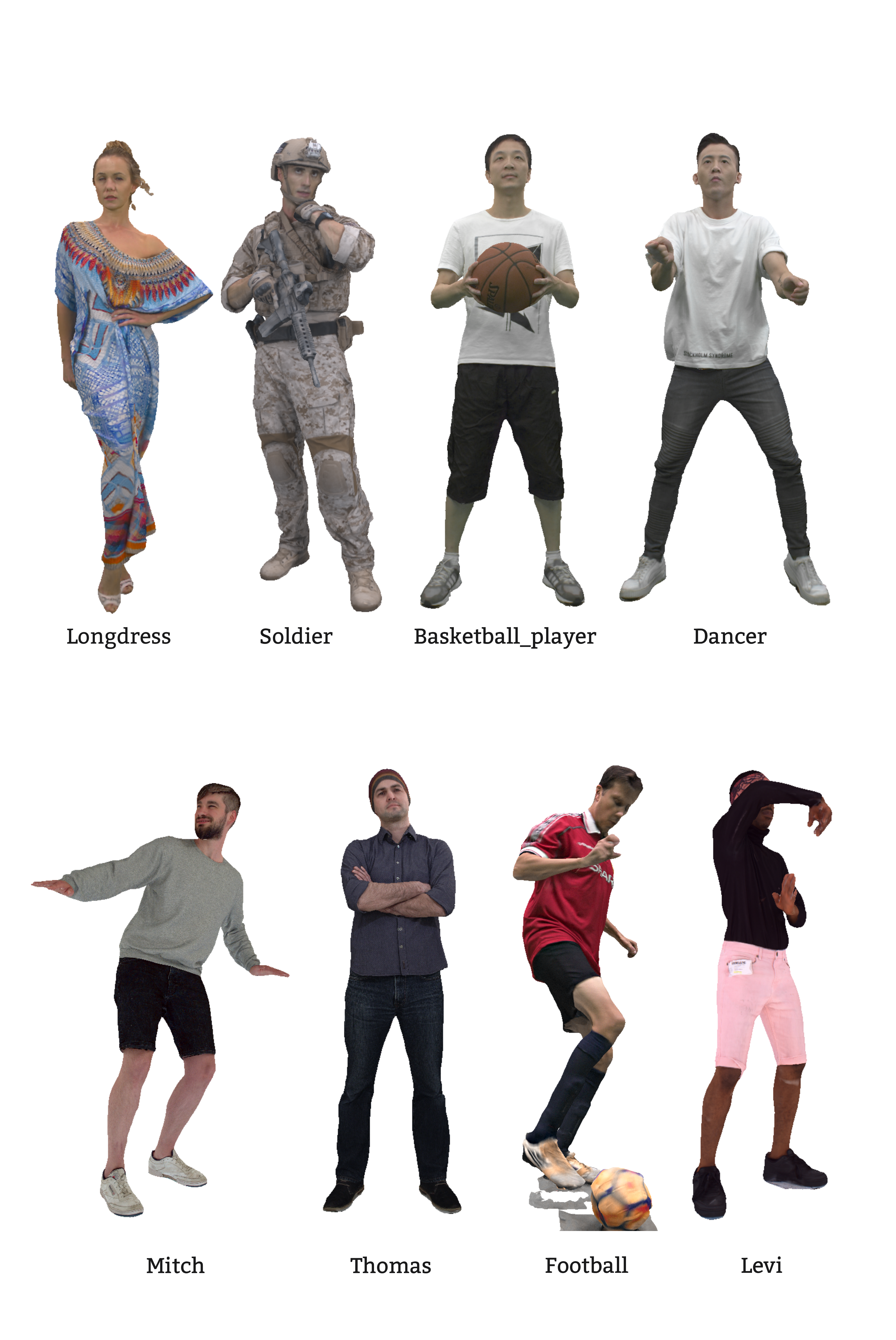}
    \caption{Snapshots of reference mesh.}
    \label{fig:database}
\end{figure}

{{For DCM, the most typical content is 3D digital human, captured by the camera array. This explains why WG 7 uses human sequences as test material to study the compression anchor \cite{MPEG-MESH-cfp}. To effectively facilitate the optimization of compression algorithms through this study, we also use the DCM sequences provided by WG 7 as reference samples.}} There are eight human samples with different characteristics: ``Longdress'' and ``Soldier'' from ~\cite{8i,8i-m}, ``Basketball\_player'' and ``Dancer'' from ~\cite{owlii}, ``Mitch'' and ``Thomas'' from ~\cite{xdprod}, ``Football'' from ~\cite{xdprod} and ``Levi'' from ~\cite{vsense}.
The eight samples are divided into three different classes according to their encoding complexity. Class A corresponds to low-precision mesh geometry and low-resolution texture map, class B corresponds to high-precision mesh geometry and low-resolution texture map, and class C corresponds to high-precision mesh geometry and high-resolution texture map. We have summarized the detailed information of the eight samples in Table \ref{Table:reference-sample}. We denote geometry precision as GP,  texture coordinate precision as TCP, and texture map size as TMS.
We illustrate each reference mesh by a snapshot in Fig. \ref{fig:database}.

\begin{table}[pt]
	\caption{Reference Dynamic Mesh Samples.  } \label{Table:reference-sample}
	\centering
	\begin{scriptsize}
  \renewcommand{\arraystretch}{1.5}
	\setlength{\tabcolsep}{0.7mm}{
		\begin{tabular}{|c|c|c|c|c|c|c|c|}
			\hline
			Class & Name & Frames & Vertices &Faces &GP & TCP &TMS   \\ \hline
			\multirow{2}{*}{A} & Longdress  & 300 &22K &40K &10 bits &12 bits &2K$\times$2K  \\ \cline{2-8}
		  & Soldier & 300 &22K &40K &10 bits &12 bits &2K$\times$2K \\ \cline{1-8}

    	   \multirow{2}{*}{B} & Basketball\_player  & 300 &20K &40K &12 bits &12 bits &2K$\times$2K  \\ \cline{2-8}
		  & Dancer & 300 &20K &40K &12 bits &12 bits &2K$\times$2K \\ \cline{1-8}

        	   \multirow{4}{*}{C} & Mitch  & 300 &16K &30K &12 bits &13 bits &4K$\times$4K  \\ \cline{2-8}
		  & Thomas & 300 &16K &30K &12 bits &13 bits &4K$\times$4K \\ \cline{2-8}

    		  & Football & 300 &25K &40K &12 bits &13 bits &4K$\times$4K \\ \cline{2-8}

        		  & Levi & 150 &20K &40K &12 bits &13 bits &4K$\times$4K\\ \cline{1-8}    
\hline
	\end{tabular}}
	\end{scriptsize}
\end{table}

\subsection{Distortion Generation} \label{sec:distortion-generation}

To explore the influence of typical mesh distortion on subjective perception, we apply six common distortion types: CN, DS, GN, MD, MLC and TC.
More specifically, we propose five or seven different distortion levels for each type of distortion, with details shown below:
\begin{enumerate}
            \item CN: Color noise is applied to the texture map of the mesh sample. This noise simulates the distortion injected into the image during channel transmission. We use the Matlab function ``imnoise'' to add ``salt \& pepper'' noise to the texture map with noise densities set to 0.01, 0.05, 0.1, 0.15, 0.2, 0.25, and 0.3.
    \item DS: Downsampling is a very simple and effective method to reduce the complexity of data processing. We use the Matlab function ``imresize'' with  cubic interpolation to resize the resolution of the texture map with ratios set to 0.5, 0.3, 0.1, 0.08, 0.05, 0.03, and 0.01, applied in each direction.
    \item GN: Geometrical Gaussian noise is applied to the spatial coordinates of the vertices of the mesh sample, to simulate the distortion injected during a serial mesh processing (e.g., mesh scanning and capture). We first generate random noise in the interval [0, 1] for each mesh vertex, then scale the noise value corresponding to the bounding box size, as shown in Eq. \eqref{eq:GN}. 
    \begin{equation}
    \label{eq:GN}
    N_{i} = NR_{i} * Ratio * MinBBlength, 
    \end{equation}
    The noise of the $i$-th vertex, $N_{i}$, is represented in Eq. \eqref{eq:GN}, where $NR_{i}$ represents the random noise in the  $i$-th vertex, $MinBBlength$ represents the minimum size of the bounding box among three dimensions. $Ratio$ controls the degree of distortion with the following values: 0.001, 0.004, 0.007, 0.01, 0.013, 0.016, and 0.02.

    \item MD: Mesh decimation is applied to the faces of the mesh sample.  The principle of mesh decimation is to reduce the number of faces and to generate models with different levels of detail for different needs. {WG 7 ~\cite{MPEG-MESH-Coding} proposes five MD degrees which are the basis of the Draco compression ~\cite{draco} to realize specific bitrates (i.e., current lossy compression anchors first conduct MD and then compress the meshes with encoder). Therefore, we directly use the MD samples proposed by WG 7 to better study the independent influence of MD. The number of faces after decimation are 25K, 20K, 15K, 10K, and 5K. Quadric edge collapse decimation [with texture] is used, as implemented by MeshLab}. For ``Levi'', MeshLab cannot perform MD, as the mesh contains several small texture map islands and the algorithm cannot achieve the target face count ~\cite{MPEG-MESH-Coding}. 
    

    \item MLC: MPEG lossy compression is applied to both the texture map and geometry information (e.g., vertex spatial coordinates and UV coordinates) of the mesh sample. It refers to the lossy compression anchor with the all-intra mode ~\cite{MPEG-MESH-Coding}. The texture map is first converted to YUV420 video format using HDRTools ~\cite{HDRtool}, and then encoded with HEVC HM 16.21 reference software with the Screen Content Coding Extension~\cite{HEVC}. For the geometry information,  MD is applied first and the information is compressed with Draco ~\cite{draco}. There are five different target bitrates (i.e., R1 to R5 with the increase of bitrate, the detailed compression parameters are presented in Appendix) for each of the eight samples proposed in ~\cite{MPEG-MESH-Coding}.  For ``Levi'', R1 cannot be reached due to the decimation problem explained above. R1 is the lowest bitrate and it requires the application of MD. Therefore, there are only four available target bitrates for Levi.


\item TC: Texture map compression is applied to study the influence of SOTA image compression algorithms. We use FFMPEG 5.1.2 with libx265 ~\cite{ffmpeg} to compress the texture map with quantization parameters (QP) set to 22, 27, 32, 37, 42, 48, and 50.

\end{enumerate}

\begin{figure}
    \centering
    \includegraphics[width=1\linewidth]{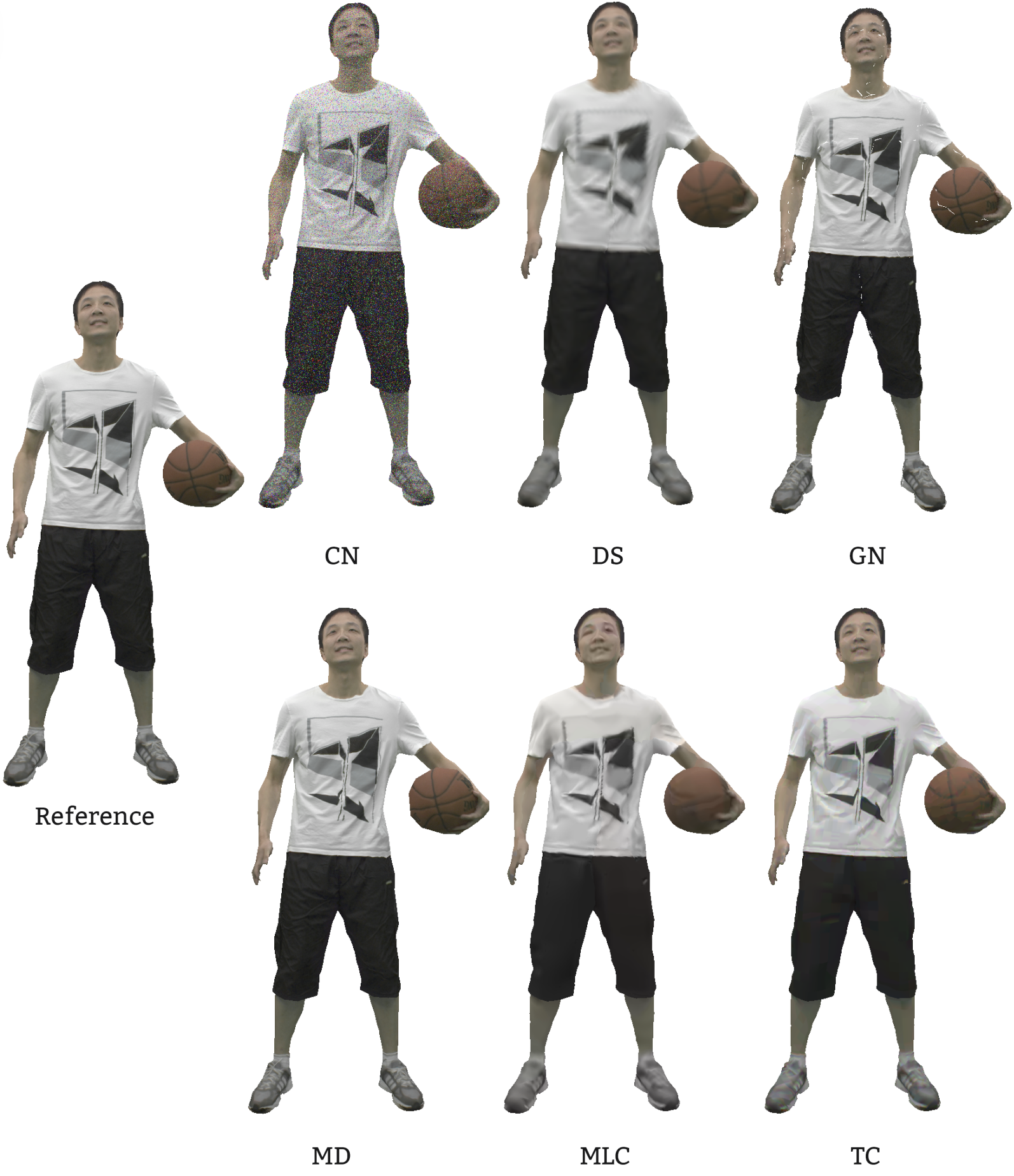}
    \caption{Snapshots of distorted mesh in TDMD.}
    \label{fig:dis-database}
\end{figure}

Fig. \ref{fig:dis-database} illustrates the snapshots of the distorted meshes for ``Basketall\_player''. All distortions considered, there are $8\times(4\times7+2\times5) - 1 = 303$ distorted samples in the proposed database (eight reference meshes, four types of distortions with seven distortion degrees, two types of distortions with five distortion degrees, and minus one for the unreachable R1 of MLC for ``Levi'').

\subsection{Video Generation}\label{sec:vg}

There are two prevalent methods to render mesh samples for subjective evaluation:  2D video-based and VR-based methods. 
For video-based subjective methods, we can refer to ITU-R BT.500 ~\cite{BT500} and ITU-T P.910 Recommendations ~\cite{11scale-rating} to conduct the subjective experiment. For VR-based methods, although there is a lot of academic research to explore the principles of the subjective experiment, an authoritative  protocol is yet to be standardized, which is part of ongoing activities in ITU-T SG12/Q7 P.IntVR~\cite{PintVR}. In this paper, we use a video-based method to render our database. 

To render the DCM samples into PVSs and fully display their geometrical and texture characteristics, we adopted specific approach to design the camera path. First, we set the views presented in Fig. \ref{fig:database} as the starting points. The camera is fixed for one second to let the viewer observe the initial view. Then, 
the camera rotates anticlockwise around the sample during four seconds before returning  to its original position, where it remains fixed again for one second. Finally,  the camera rotates clockwise around the sample to return to its original position within four seconds. 
 Overall, a DCM sample of 300 frames is converted to a smooth 10-second video at 30 frames per second.  For Levi, which only has 150 frames, we use the inverse order of the original 150 frames, indexed as the 151${}^{\rm st}$ to  300${}^{\rm th}$ frames, to obtain the 10-second videos with the camera path defined above. 

We use Open3D ~\cite{open3d}  to render the mesh samples and capture the images at a resolution of $1920\times1080$. After obtaining the images, we use FFMPEG ~\cite{ffmpeg} to group them into videos with libx265. The constant rate factor (CRF) is set to 10, which guarantees smooth display and visually lossless degradation, as suggested in ~\cite{m57896}.

\subsection{Subjective Test Methodology}
In this part, we introduce the design of our subjective experiment, including stimulus and scale, the setting of the training and rating sessions, as well as the test environment.
\subsubsection{Stimulus and Scale}
We use the double stimulus impairment scale (DSIS) where the reference PVS is first displayed, followed by two-second breaks with a white blank screen. The distorted PVS is then displayed, and followed by a white blank screen for eight seconds, allowing the viewer to rate  the PVS. For rating samples, we use the 11-grade voting method proposed by ITU-T P.910 Recommendation. 


\subsubsection{Training Session}
To collect reliable subjective scores from viewers and reduce outliers, a clear and judicious training session is needed to help viewers foster a good understanding of the relationship between mesh quality and subjective scores. 




 
We carefully select eight distorted PVSs for the training session, corresponding to the predefined expected quality range. The predefined expected quality scores are evenly distributed between the lowest score and the highest score. The training session is displayed twice, and the viewers are required to score the eight distorted PVSs in the second round. If the scores given by the viewer present a high correlation with the expected quality interval, it is considered that the viewer has understood the principle of the subjective experiment and is qualified for the rating sessions. Otherwise, we repeat the training procedure until the viewer can give reasonable scores for the training PVSs, without ever telling the viewer how to score a certain training PVS.
 

\subsubsection{Rating Session}
\begin{figure}[h]
\vspace{-1.5em}
    \centering
    \includegraphics[width=1\linewidth]{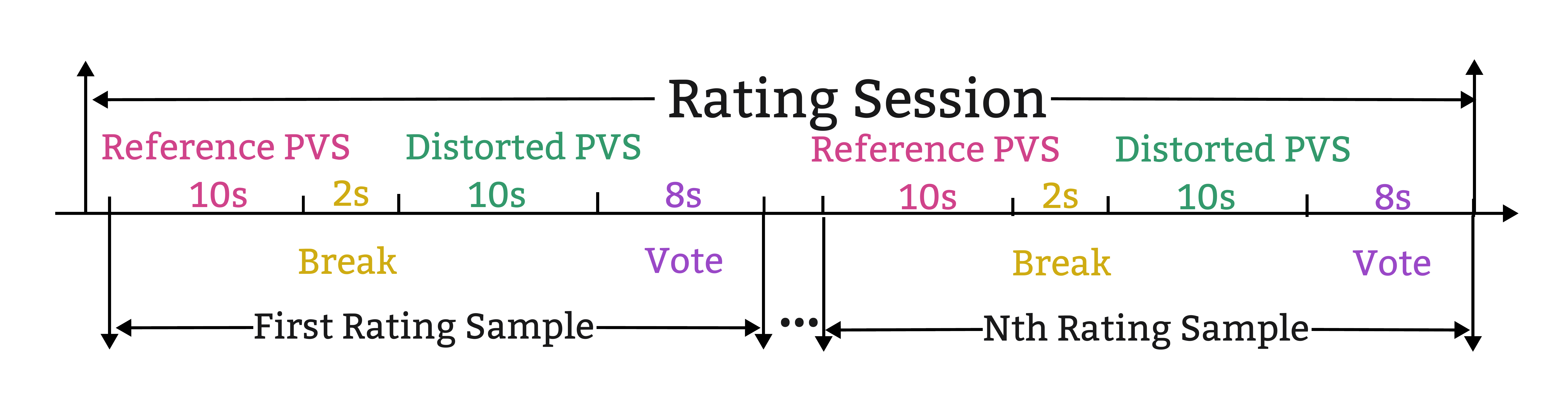}
    \caption{Procedure of rating session. $s$ represents second (s).}
    \label{fig:testsession}
\end{figure}

 The rating procedure is shown in Fig. \ref{fig:testsession}: 30s are spent on each distorted sample. As mentioned in section \ref{sec:distortion-generation}, there are 303 distorted samples to rate. To avoid visual vertigo and fatigue caused by an overly long experiment time, we limit the duration of a single rating session to 20-30 minutes by randomly splitting the 303 distorted samples into 12 subgroups, having 25 to 30 samples each.

\subsubsection{Test Environment}
PVSs are displayed in their original resolution on a 4K monitor (PHILIPS 329P1, 32 inches with 3840$\times$2160 resolution) in a low-light lab environment. We recruited 63 people for subjective experiments, 35 females and 28 males aged from 18 to 50 years old.  They are company employees or university students. Most of them have no experience with subjective experiments, and are naive to the DCM processing activity.  Each viewer is only presented  some of the subgroups since going through the entire database is a relatively onerous task. When a viewer participates in the scoring of more than two subgroups, we ensure that there is a sufficient interval time between the first and the second subjective experiments to avoid cross-influence. We collect at least 25 groups of raw scores for each subgroup to ensure that we will have at least 16 valid scores after the outlier removal.

\subsection{Outlier Detection}\label{sec:the proposed method}
We apply two consecutive steps to remove outliers from the raw subjective scores. In the first step, we insert ``trapping samples'' for each subgroup. Two types of trapping samples are considered. Firstly, we add a very low quality PVS, expected to be scored 0, 1 or 2.  If a viewer gives a higher score for this sample, the scores of this subgroup collected from this viewer are considered incorrect. Secondly, we add a randomly selected duplicate PVS in each subgroup. These PVSs are not displayed consecutively. The score differences of the duplicated PVSs should be below 3, otherwise, the scores of this subgroup collected from this viewer are considered wrong. In the second step, we calculate the average score for all the samples in the database, then we calculate the correlation between the average score and the raw scores of each viewer via Eq. \eqref{eq:correl},
    \begin{equation}
    \label{eq:correl}
    \rho_{x,y} = \frac{Cov(x,y)}{\sigma_{x}\times\sigma_{y}},
    \end{equation}
where the correlation of the array $x$ and $y$, $\rho_{x,y}$, is calculated based on the covariance $Cov(x,y)$ and the variances $\sigma_{x}$, $\sigma_{y}$. We remove the scores of viewers whose correlation is lower than 0.8 and then update the average scores to get the MOS. We do not adopt the outlier detecting method proposed in ITU-R BT.500 as our viewers have only rated  parts of the database, while the ITU-R BT.500 method requires each viewer to rate the entire database.

Six groups of scores {(scores from a rating session are labeled as a group of scores)} were detected as dubious scores in the first step, and the scores of one viewer were removed in the second step.
\subsection{MOS Analysis}\label{sec:mos-ana}
To illustrate the influence of different distortions on human perception, we plot in Fig. \ref{fig:mos_curve} the MOS distribution versus the distortion types. We can observe that: 
\begin{figure}[h]
	\centering
	\subfigure{
		\label{fig2.1}
		\includegraphics[width=0.45\linewidth]{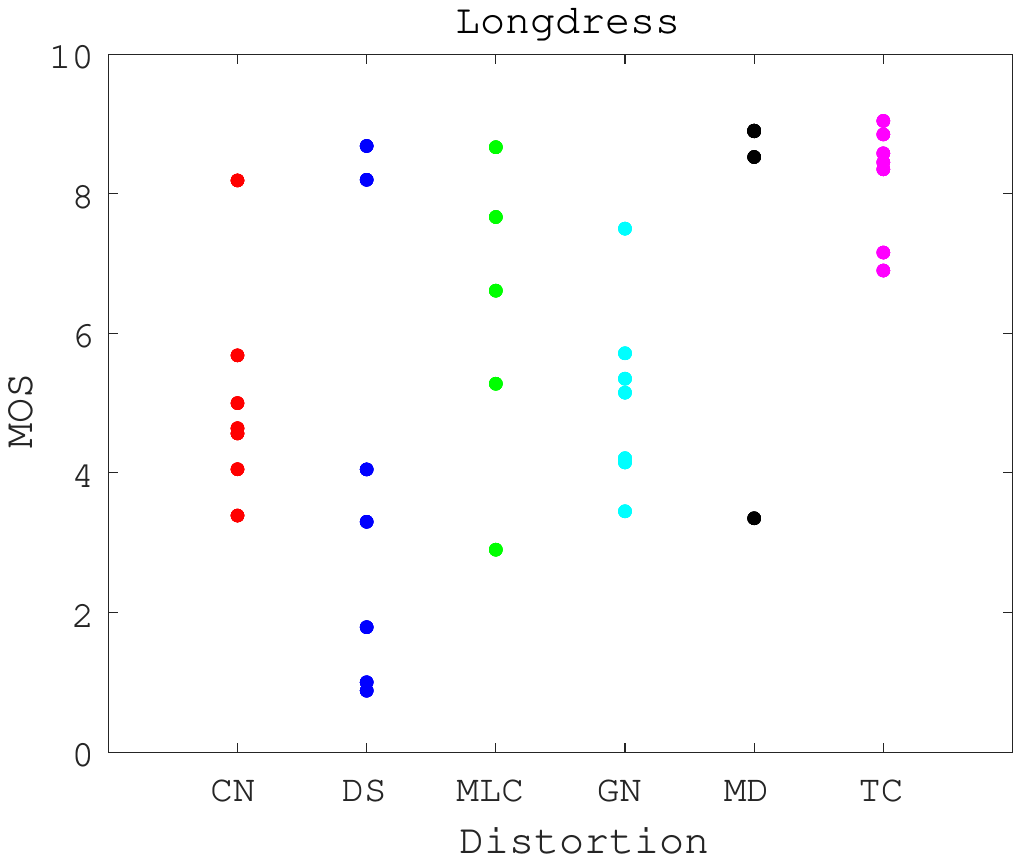}}
	\subfigure{
		\label{fig2.2}		\includegraphics[width=0.45\linewidth]{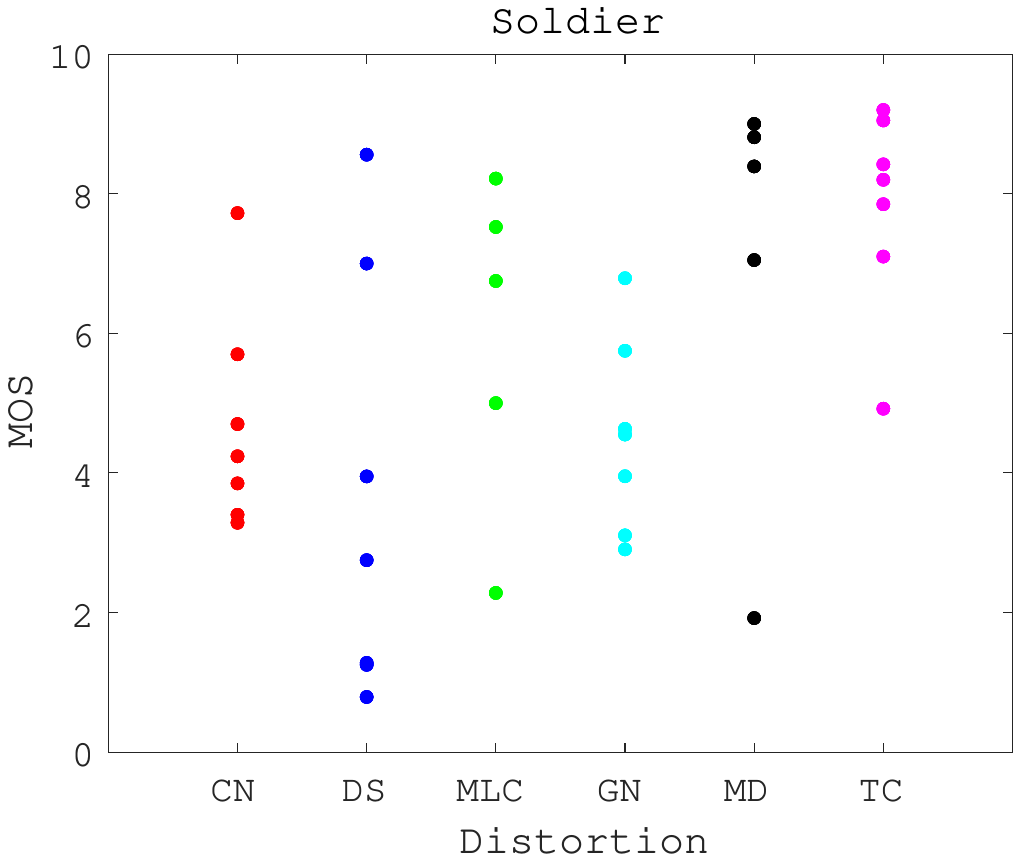}}\\
	\subfigure{
		\label{fig2.3}
		\includegraphics[width=0.45\linewidth]{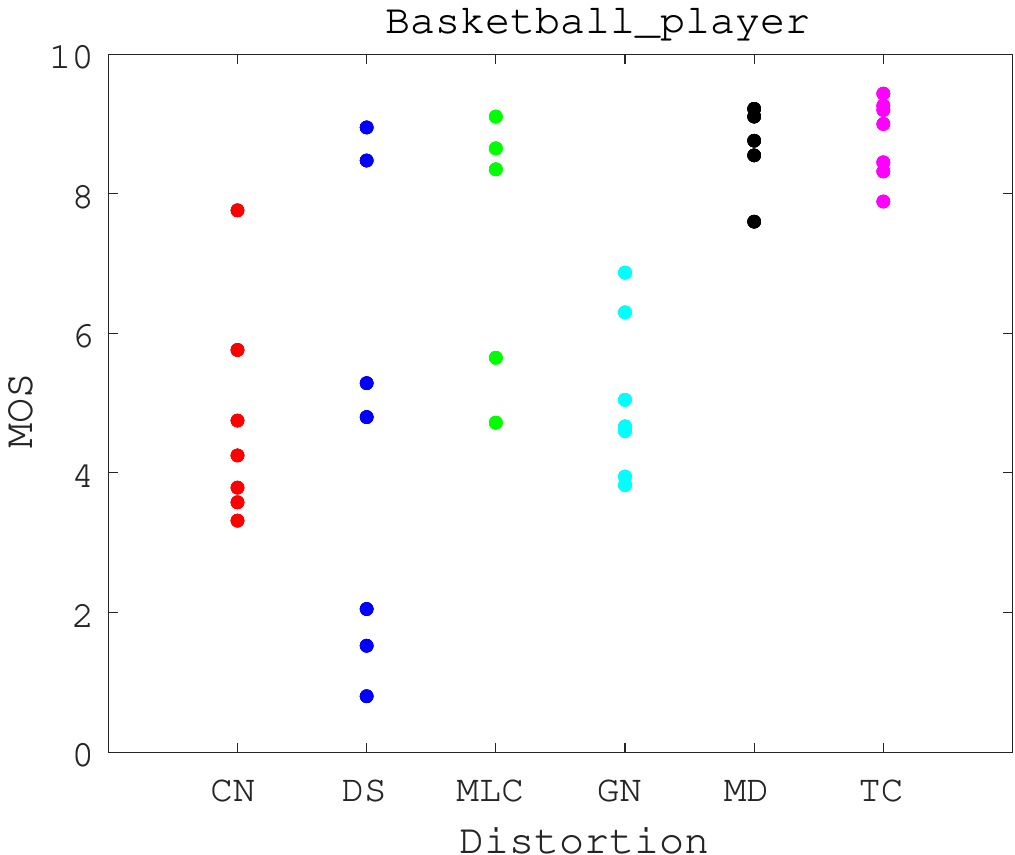}}	
	\subfigure{
		\label{fig2.4}
		\includegraphics[width=0.45\linewidth]{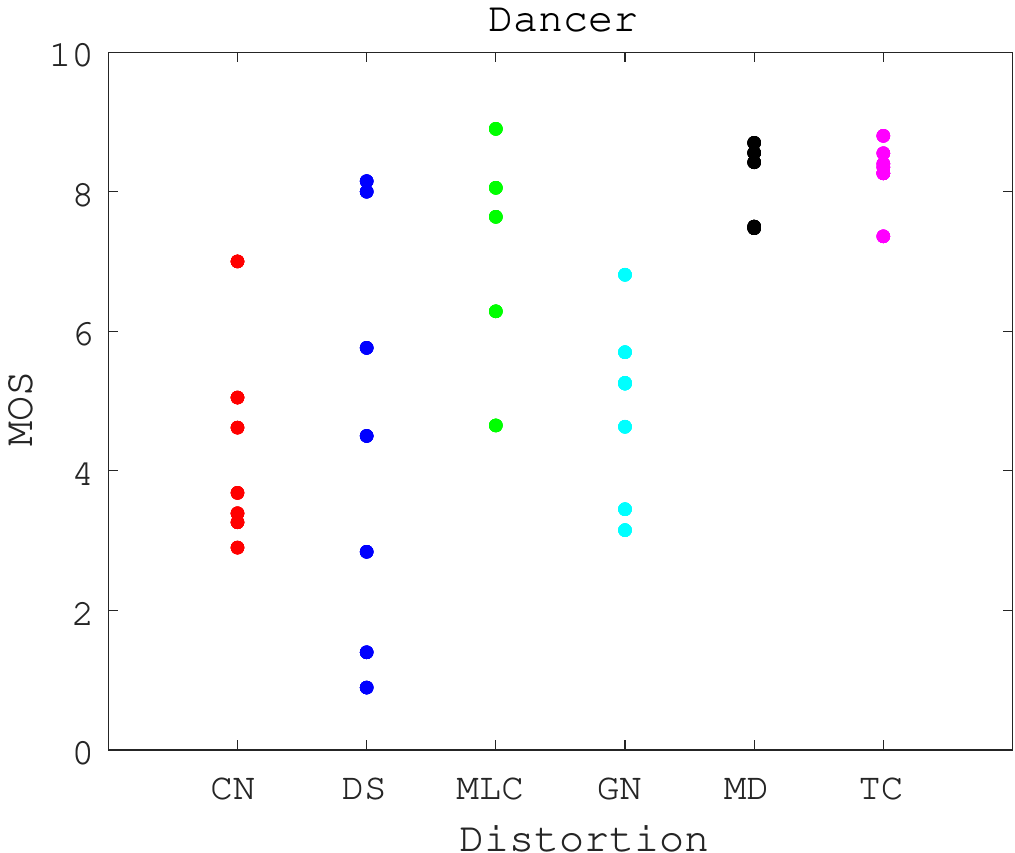}}\\		
	\subfigure{
		\label{fig2.5}
		\includegraphics[width=0.45\linewidth]{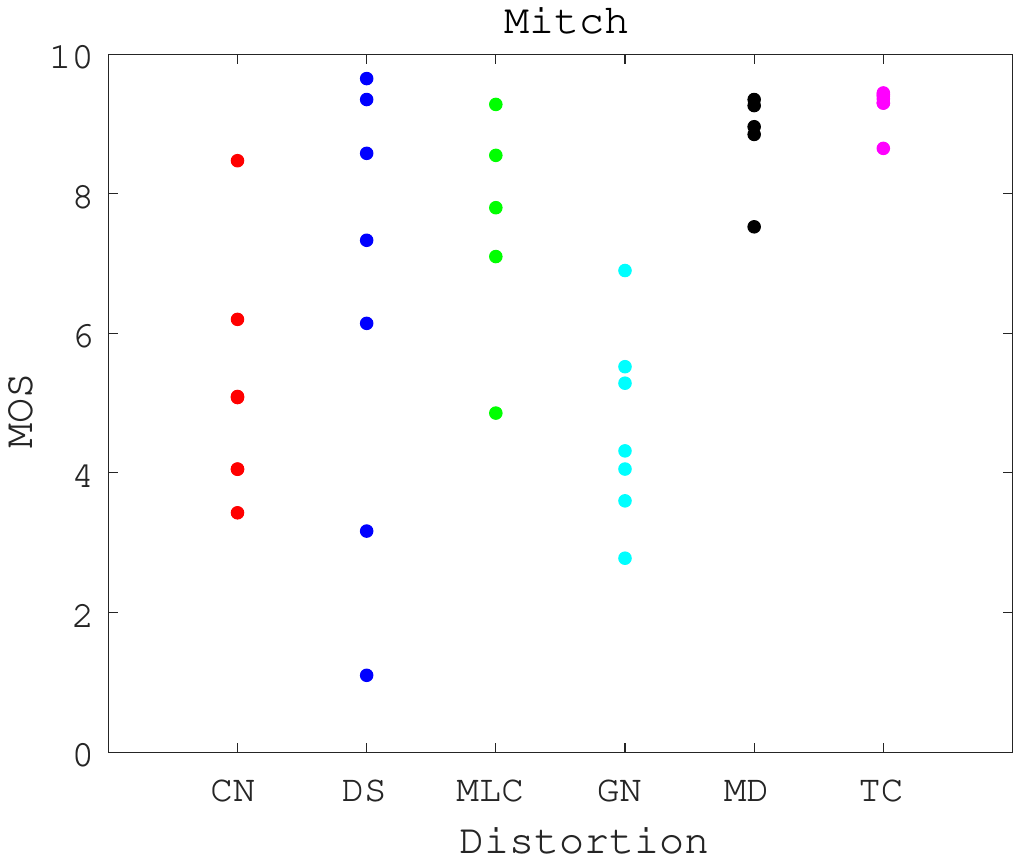}}
	\subfigure{
		\label{fig2.6}
		\includegraphics[width=0.45\linewidth]{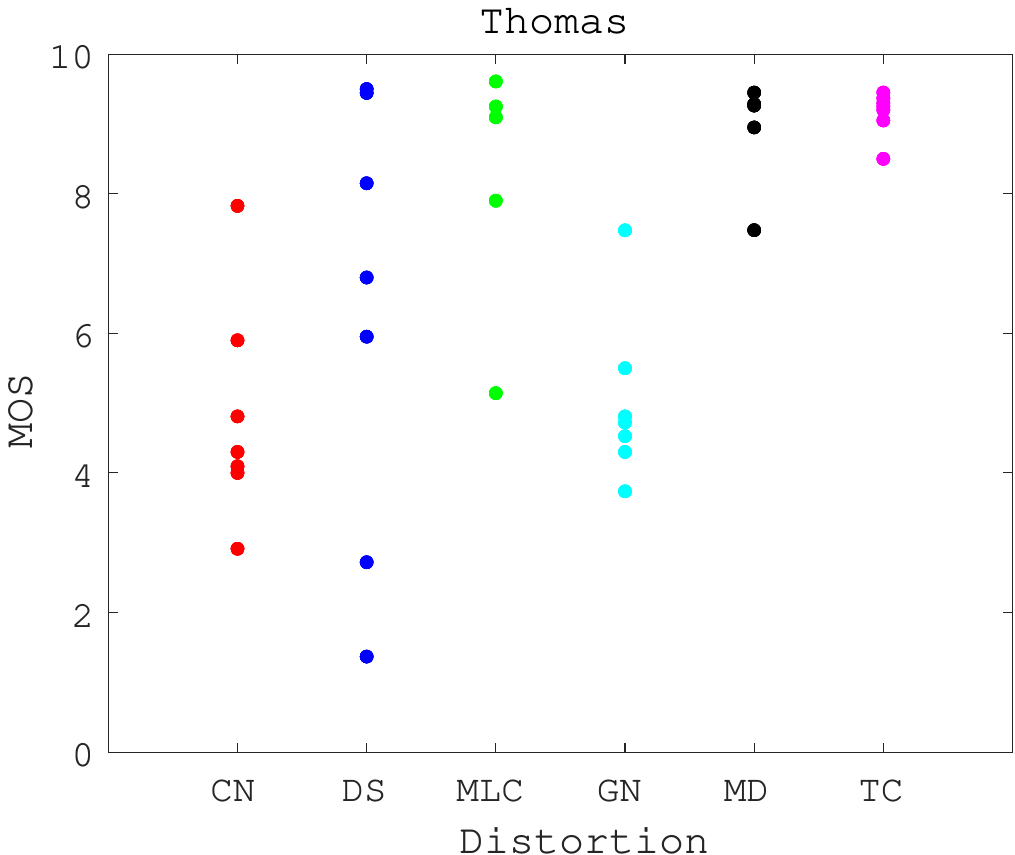}}\\
	\subfigure{
		\label{fig2.7}
		\includegraphics[width=0.45\linewidth]{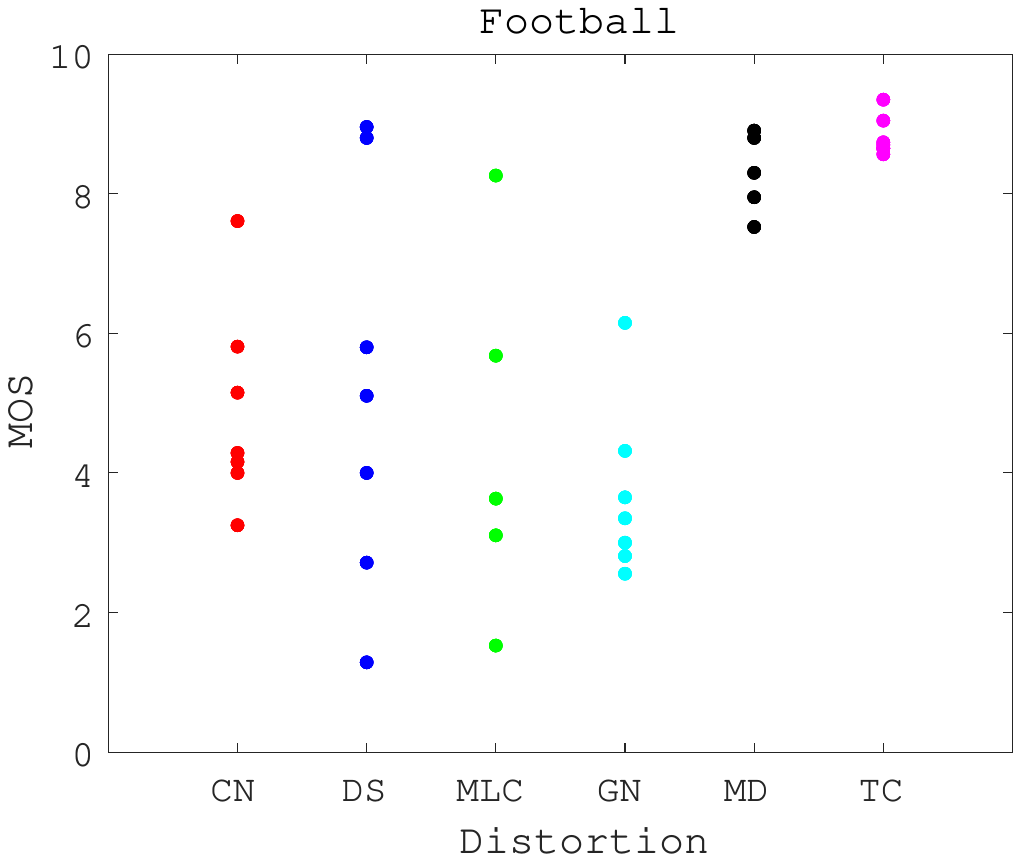}}
	\subfigure{
		\label{fig2.8}
		\includegraphics[width=0.45\linewidth]{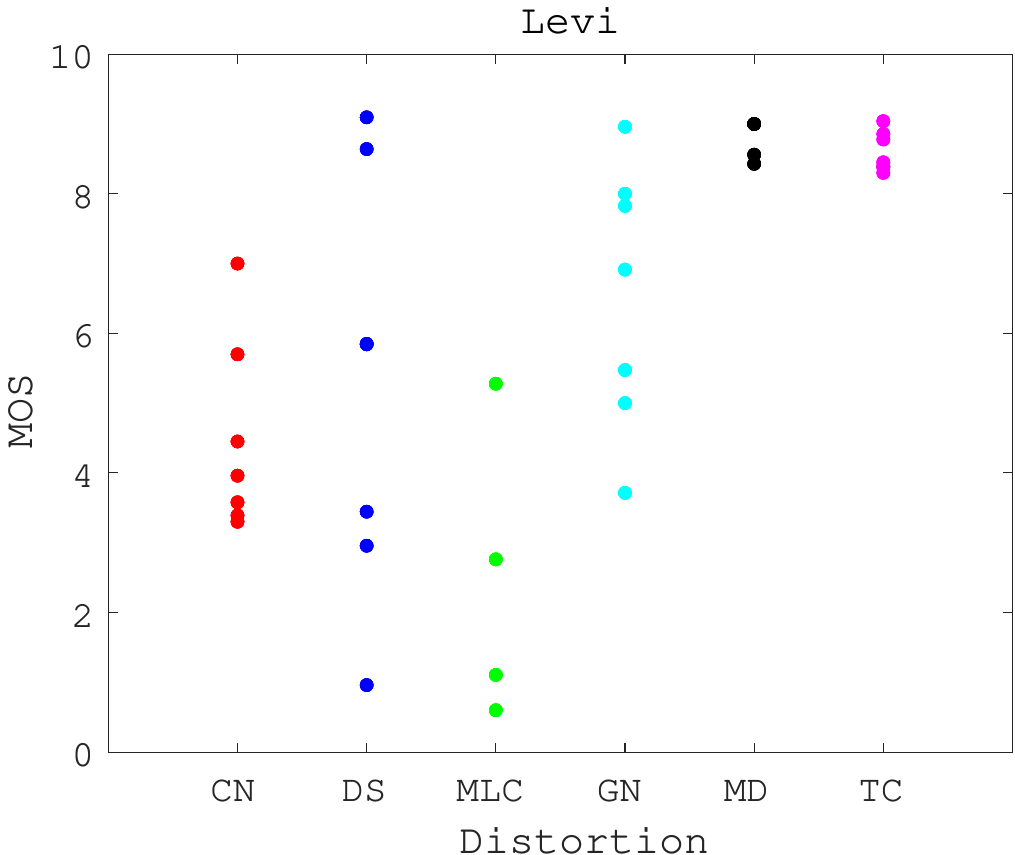}}	
	\caption{MOS plots for the $8$ test sequence.}
	\label{fig:mos_curve}
\end{figure}

\begin{enumerate}
    \item For CN, DS, MLC, and GN distortions, the MOSs cover wide value ranges. For instance, the MOS of CN for ``Longdress'' ranges from 3 to 8, and the MOS of DS for ``Soldier'' ranges from 1 to 9.
    \item For MD and TC distortions, the MOS values are generally high, most of which are higher than 7. 
    It indicates that the influence of MD and TC distortion degree considered in this paper on perceptual quality is limited. 
\end{enumerate}

Based on the above experimental results, the MOS variations of CN, DS, MLC, and GN, which present wide quality variation, satisfy our expectations. We consequently focus our analysis on studying why the degree increase of MD and TC has a limited impact on human perception.


By carefully checking MD samples, we postulate that the main reason for having limited perceptual impact is that the MD degrees proposed by WG 7 are low, resulting in a number of faces after MD that remain large enough to preserve most sample characteristics. For these MD samples, contour flicker is the main manifestation of perception quality degradation, which is less distracting than other types of distortion.




By carefully checking TC samples, we hypothesize that the main reason having a limited impact is that the noticeable distortions caused by TC are relatively small, because they are masked by the frame switching of the dynamic sequence. The detailed analysis of MD and TC is presented in the Appendix.


\section{Objective DCM Metrics}\label{sec:obj-me-re}
To evaluate the quality of DCM, WG 7\cite{MPEG-MESH-metric} proposes two methods: an evaluation via image-based metrics using conventional mesh rendering techniques and an evaluation via point-based metrics using sampling to convert the mesh into a point cloud. Another method, labeled as the video-based metrics, that infers mesh quality by applying IQA/VQA metrics on the PVSs rendered in the subjective experiment.  In this section, we first give a brief introduction to the three types of metrics before analyzing their performance on the proposed database. Additionally, we explore the influence of sampling on SOTA point-based metrics.

\subsection{Description of the Metrics }
\subsubsection{Image-based Metrics}
Before applying image-based metrics, we must first convert reference and distorted meshes into images. We do this by projecting them through 16 viewpoints resulting from the Fibonacci sphere lattice ~\cite{fibonacci}.


Based on the 16 obtained images, we tested three metrics based on the proposal in ~\cite{MPEG-MESH-metric}: $\rm geo_{psnr}$, $\rm rgb_{psnr}$ and $\rm yuv_{psnr}$. $\rm geo_{psnr}$ calculates the depth information differences between the reference and the distorted meshes. The depth information is recorded when capturing the images as an image itself, whose pixel values are normalized to 255, to get PSNR values comparable to the ones obtained using the next two metrics.  $\rm rgb_{psnr}$ and $\rm yuv_{psnr}$ calculate the differences of $(R, G, B)$ and $(Y, U, V)$ color channels between the reference and the distorted images. When capturing the images, the color information is recorded as an image. A more detailed description of the metrics can be found in ~\cite{MPEG-MESH-metric}. 

The frame rates of DCM sequences are usually 30Hz ~\cite{MPEG-MESH-Coding}, and the videos we generated have a frame rate of 30Hz as well. To balance accuracy and computation complexity, we propose to calculate the objective scores at a frequency of three frames per second. Based on our testing, the results are close between using all the frames and using three frames per second.  Therefore, to obtain an overall objective score for a DCM that contains hundreds of frames, we calculate the objective scores every ten frames, which are then averaged to get the final results.  

\begin{table*}[pt]
	\caption{Metrics performance on TDMD. } \label{Table:Metric-perfromance-1}
	\centering
	\begin{scriptsize}
 \renewcommand{\arraystretch}{1.5}
	\setlength{\tabcolsep}{0.7mm}{
		\begin{tabular}{|c|c|c|c|c|c|c|c|c|c|c|c|c|c|c|c|c|c|}
			\hline
			 \multicolumn{2}{|c|}{Index}& \multicolumn{2}{c|}{All} & \multicolumn{8}{c|}{Sequence}&\multicolumn{6}{c|}{Distortion}  \\ \hline
			\multicolumn{2}{|c|}{Metric}&PLCC&SROCC&Longdress&Soldier&Basketball\_player&Dancer&Mitch&Thomas&Football&Levi& CN&DS&GN&MLC&MD&TC \\  \hline		

    \multirow{3}{*}{$\bf A$}&$\rm geo_{psnr}$& 	0.48 & 	0.16 &0.48& 	0.52& 	0.14& 	0.52& 	0.54&0.54& 	0.21& 	0.37 &-& 	-& 	0.82& 	0.76& 	0.85& 	- \\ \cline{2-18}
    &$\rm rgb_{psnr}$& 	0.85&	0.83& 0.91 	&0.89& 	0.90& 	0.92& 	0.93& 	0.92& 	0.89& 	0.84
     & 0.96& 	0.91& 	0.83& 	0.80& 	0.81& 	0.56\\ \cline{2-18}
    &$\rm yuv_{psnr}$& 	0.86& 	0.84 &0.91& 	0.90& 	0.92& 	0.92& 	0.93& 	0.93& 	0.89& 	0.84 &0.96& 	0.92 &	0.85 &	0.80 &	0.80& 	0.53 \\ \hline 
 \hline
\multirow{6}{*}{$\bf B$}& D1&	0.49&	0.16& 0.47& 	0.51 &	0.50& 	0.50 &	0.55 &	0.53 	&0.48 	&0.57  &-&	-&	0.77 &	0.89 &	0.96 &	-\\ \cline{2-18}

&D2 &0.50&0.16&0.47 &	0.51 &	0.50 &	0.50 &	0.55 &	0.54 &	0.49 &	0.58 	&-&	-	&0.79 &	0.90 &	0.96 &	- \\ \cline{2-18}

&D1-h &	0.43&	0.13&0.40 &	0.46 &	0.45 &	0.48 &	0.47 &	0.43 	&0.31 	&0.55  &-	&-	&0.79 &	0.92 &	0.94 	&-\\ \cline{2-18}

&D2-h&	0.45&0.14& 0.45 &	0.49 &	0.46 	&0.46 	&0.49 	&0.48 &	0.42& 	0.53 &-	&-&	0.79 &	0.88 &	0.95 &	-\\ \cline{2-18}

&$\rm yuv_{p}$&	0.82&	0.80&0.91 &	0.85 &	0.93 	&0.93 	&0.93& 	0.93 &	0.91 &	0.94 &0.96 &	0.82 &	0.78 &	0.86 &	0.52 &	0.41\\ \cline{2-18}

& ${\rm PCQM}_{\rm p}$&0.91& 	0.87&0.93 	&0.93 &	0.95 	&0.95 	&0.95 &	0.92 	&0.93& 	0.94  &0.96 &	0.95 &	0.88 &	0.83 &	0.92 &	0.66 	

  \\ \hline
\hline
\multirow{6}{*}{$\bf C$}& PSNR&0.78  &	0.78& 0.80 &	0.71 &	0.76 &	0.79 &	0.85 	&0.73 &	0.82 &	0.88 &0.96 &	0.91 &	0.86 &	0.94 &	0.58 &	0.67	 \\ \cline{2-18}
& SSIM &0.84  	&0.81& 0.91 &	0.90 &	0.86 	&0.87 &	0.90 &	0.88 &	0.85 &	0.94 &0.93 &	0.90 &	0.92 &	0.88 &	0.93 &	0.48	  \\ \cline{2-18}
& MS-SSIM &	0.90 &	0.88&0.94 &	0.94 	&0.91 &	0.92 &	0.93 &	0.91 &0.91& 	0.97&0.94 &	0.95 &	0.93 &	0.93 &	0.93 &	0.74  \\ \cline{2-18}
& VMAF & 	0.80  &	0.79&0.82 &	0.77 &	0.80& 	0.79 &	0.87 &	0.77 &	0.85 &	0.86 &0.95 	&0.94 &	0.86 &	0.94 &	0.88 &	0.82 \\ \cline{2-18}
&VQM &	0.79&	0.77& 0.82 &	0.78 	&0.83 &	0.83 &	0.91 &	0.84 &	0.83 &	0.75 & 0.95 &	0.91 &	0.83 &	0.76 &	0.80 &	0.69 \\ \cline{2-18}
&3SSIM &	0.83  &	0.81 & 0.88 &	0.85 	&0.87 &	0.86 &	0.86 &	0.84 &	0.88 &	0.94&0.93 &	0.84 &	0.85 	&0.94 &	0.93 &	0.71   \\ \hline
	\end{tabular}}
	\end{scriptsize}

\end{table*}

\subsubsection{Point-based Metrics}

The point-based metrics directly use the raw data from the reference and the distorted meshes to extract features and predict quality. Two steps are applied ~\cite{MPEG-MESH-metric}. Firstly, meshes are sampled to be converted into point clouds. Secondly, point cloud objective metrics are calculated to predict quality. Several methods for sampling have been proposed, including grid sampling, face sampling, and surface subdivision sampling. As ~\cite{MPEG-MESH-metric} reported that grid sampling has a stable behavior, it is used in this section with a grid resolution of 1024 to generate colored point clouds. 
This grid resolution controls the point density of the sampled point cloud: the larger the resolution, the denser the point cloud. 
More results corresponding to other types of sampling and values of resolution are presented in section \ref{sec: exp_sampling}.

After generating the point clouds, six SOTA point cloud metrics are computed: p2point (D1), p2plane (D2), p2point-hausdorff (D1-h), p2plane-hausdorff (D2-h), $\rm yuv_{psnr}$ ($\rm yuv_{p}$) and ${\rm PCQM}_{psnr}$ (${\rm PCQM}_{\rm p}$). In the same manner as for image-based metrics, we calculate the objective scores every ten frames and average them to obtain the final results. 

\subsubsection{Video-based Metrics}
To apply IQA/VQA metrics, we use the PVSs displayed during the subjective experiment. The MSU video quality metric benchmark ~\cite{antsiferova2022video} is used to compute the following metrics: PSNR, SSIM ~\cite{wang2004image}, MS-SSIM ~\cite{wang2003multiscale}, VMAF ~\cite{vmaf-li2016toward}, VQM~\cite{vqm} and 3SSIM ~\cite{li2009three}.

\subsection{Performance of the Metrics}
To ensure the consistency between the objective and subjective evaluation scores for the various quality assessment metrics, the Video Quality Experts Group recommends mapping the dynamic range of the scores from objective quality assessment metrics to a common scale using five-parameters logistic regression ~\cite{video2003final}, e.g., 
\begin{equation}\label{eq:log}
  Q_{i}= k_{1}\left (\frac{1}{2}-\frac{1}{1+e^{k_{2}(s_{i}- k_{3})}} \right )+k_{4}s_{i}+ k_{5},
\end{equation}
where $s_{i}$ is the objective score of the $i$-th distorted samples, $Q_{i}$ is the corresponding  mapped score. $k_1$, $k_{2}$, $k_{3}$, $k_4$ and $k_5$ are the regression model parameters to be determined by minimizing the sum of squared differences between the objective and subjective evaluation scores. 

Besides, two performance correlation indicators commonly used by the quality assessment community are employed to quantify the efficiency of the object metrics: PLCC and SROCC. PLCC can reflect prediction accuracy, and SROCC can reflect prediction monotonicity. The higher the values of the PLCC or SROCC, the better the performance of the metric ~\cite{yang2020msea}.

Table \ref{Table:Metric-perfromance-1} shows the results of the objective metrics applied on TDMD. In addition to the PLCC and SROCC results on the whole dataset, marked as ``All'', we report the PLCC results for different sequences and different distortion types in ``Sequence'' and ``Distortion'' columns. '- means that the results of the metric for the samples applied with this kind of distortion are meaningless, and the reasons will be explained in the following analysis.

\subsubsection{Image-based Metric}
Table \ref{Table:Metric-perfromance-1}-$\bf A$ presents the performance of image-based metrics. We observe that the $\rm geo_{psnr}$ results are significantly inferior than the $\rm rgb_{psnr}$ and $\rm yuv_{psnr}$ results, with an overall PLCC and SROCC of only 0.48 and 0.16, respectively. $\rm rgb_{psnr}$ and $\rm yuv_{psnr}$ have a close performance, with average PLCC and SROCC around 0.85. $\rm rgb_{psnr}$ reports a slightly more stable results than $\rm yuv_{psnr}$, the standard deviations of $\rm rgb_{psnr}$ and $\rm yuv_{psnr}$ with respect to different sequences are 0.028 and 0.030, respectively.

For the performance of image-based metrics on different distortions,  we see that $\rm geo_{psnr}$ presents high PLCC for GN, TMC, and MD distortions. However, it cannot detect CN, DS and TC distortions, as it only considers depth information which belongs to geometrical feature, while CN, DS, and TC are lossless with regards to geometry information. Therefore, $\rm geo_{psnr}$ reports the same objective scores for these samples.  $\rm rgb_{psnr}$ and $\rm yuv_{psnr}$ show close performance in different types of distortions. They exhibit high PLCC for CN and DS distortions (above 0.9) and low correlations for TC distortions (below 0.6). 

In summary, projecting DCM into colored images and then applying metrics such as $\rm rgb_{psnr}$ and $\rm yuv_{psnr}$ is more effective than only capturing depth information to apply $\rm geo_{psnr}$. One influencing factor of image-based metrics is the number of viewpoints used to generate images. With the increase in the number of projected images, the performance of image-based metrics becomes more stable~\cite{yang2020predicting,alexiou2019exploiting,icme2023}, but this is accompanied by a higher calculation complexity.

\subsubsection{Point-based Metric}\label{sec:point-based-performance}

Table \ref{Table:Metric-perfromance-1}-$\bf B$ presents the performance of point-based metrics. We see that D1 and D2 show close performance, with average PLCC and SROCC values of around 0.5 and 0.15. D1-h and D2-h exhibit close results, with overall PLCC and SROCC around 0.45 and 0.13. $\rm yuv_{p}$ demonstrates clearly higher PLCC and SROCC than D1 (-h)/D2 (-h), with average PLCC and SROCC of 0.82 and 0.80. ${\rm PCQM}_{\rm p}$ reports the best performance among all point-based metrics, showcasing overall PLCC and SROCC values of 0.91 and 0.87. The results of ${\rm PCQM}_{\rm p}$ are more stable than $\rm yuv_{p}$: the standard deviation of $\rm yuv_{p}$ between different sequences is 0.029, while that of ${\rm PCQM}_{\rm p}$ is 0.012.

For the performance of point-based metrics on different distortions, we see that D1 (-h) and D2 (-h) reveal close results on GN, MLC and MD distortions, but cannot handle CN, DS and TC distortions, the reason being that these four metrics only consider geometry information when quantifying distortion, while CN, DS and TC are color-related distortions. $\rm yuv_{p}$ shows unstable performance among the different types of distortions. It displays high correlations for CN distortions (above 0.90), but low correlations for MD and TC distortions (below 0.60). Considering that the features used in $\rm yuv_{p}$ are point-to-point differences on the Y, U, and V channels, it is more suitable for point-wise color distortion than other metrics. ${\rm PCQM}_{\rm p}$ demonstrates nearly the best and most stable results on different distortions. ${\rm PCQM}_{\rm p}$ pools local surface curvature and color features together and, therefore, showcases robust performance for all types of distortions. 

In summary, partial point-based metrics (e.g., ${\rm PCQM}_{\rm p}$) demonstrate impressive performance on the proposed database, and using both geometry and color information can significantly improve their effectiveness and robustness.

\subsubsection{Video-based Metric}
\begin{figure}[h]
\setlength{\abovecaptionskip}{0.cm}
\setlength{\belowcaptionskip}{-0.cm}
	\centering
	\vspace{-0.35cm}
	\subfigure{
		\label{fig1.1}
		\includegraphics[width=0.48\linewidth]{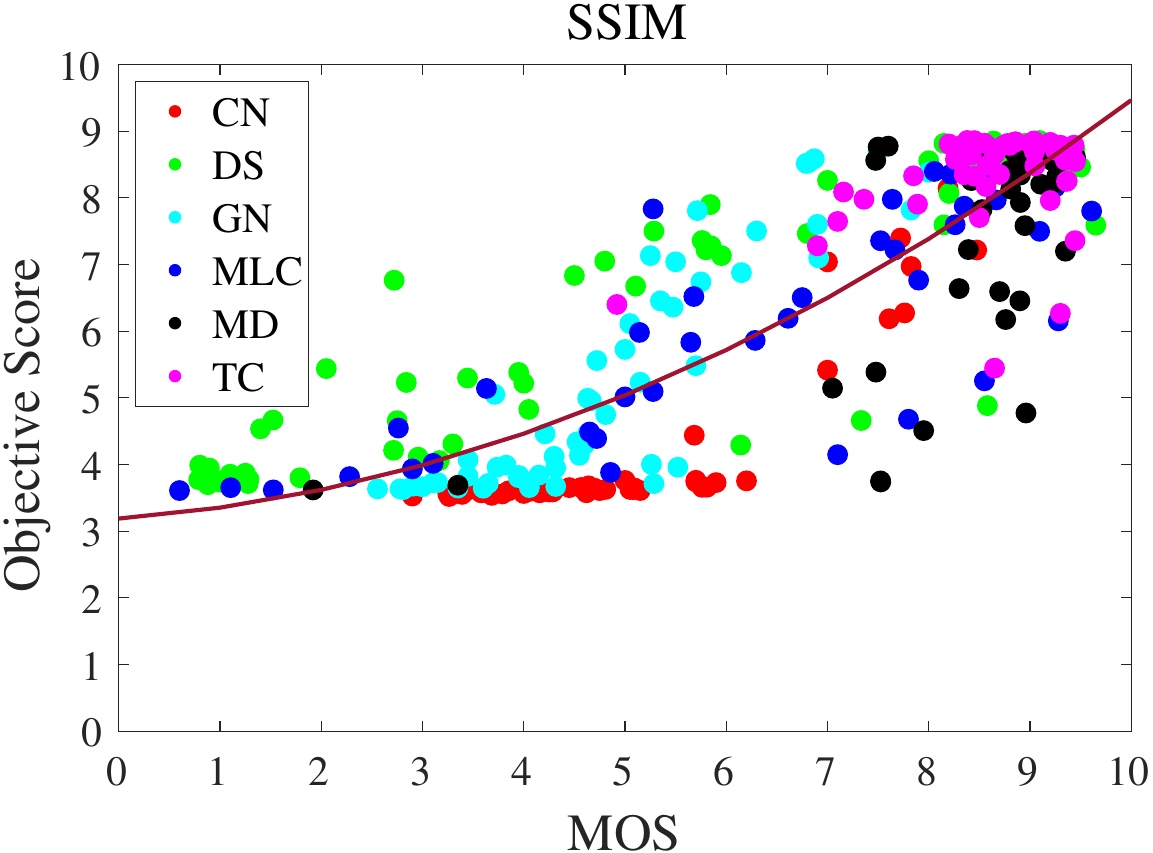}}
	\subfigure{
		\label{fig1.2}
  \includegraphics[width=0.48\linewidth]{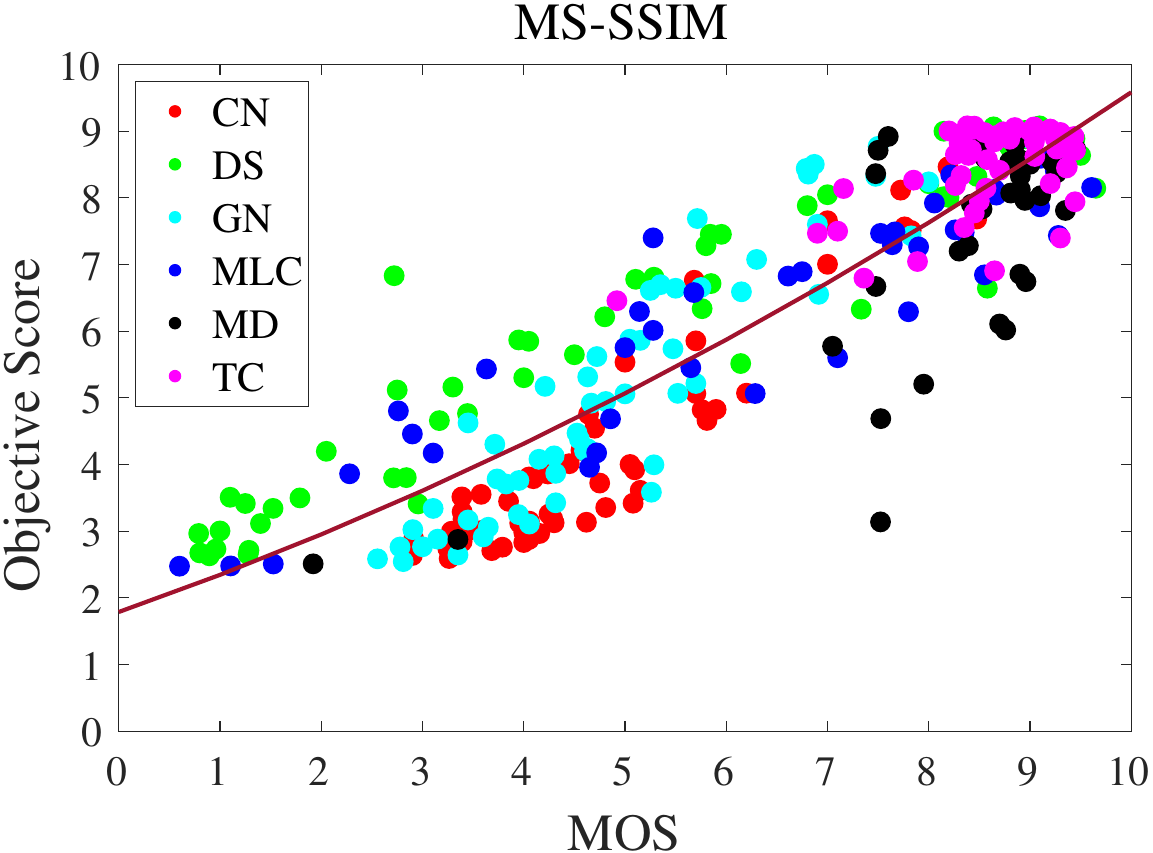}}\\ 
	\subfigure{
		\label{fig1.3}
		\includegraphics[width=0.48\linewidth]{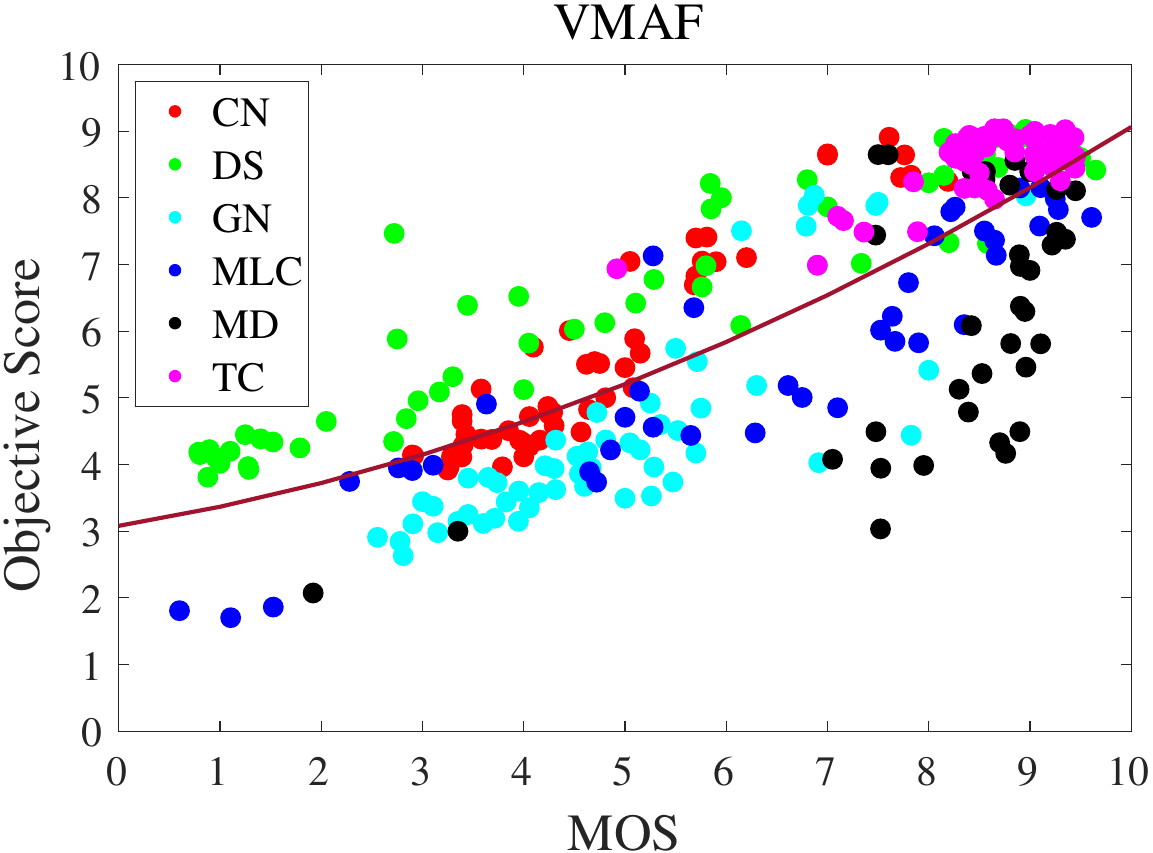}}
	\subfigure{
		\label{fig1.4}
		\includegraphics[width=0.48\linewidth]{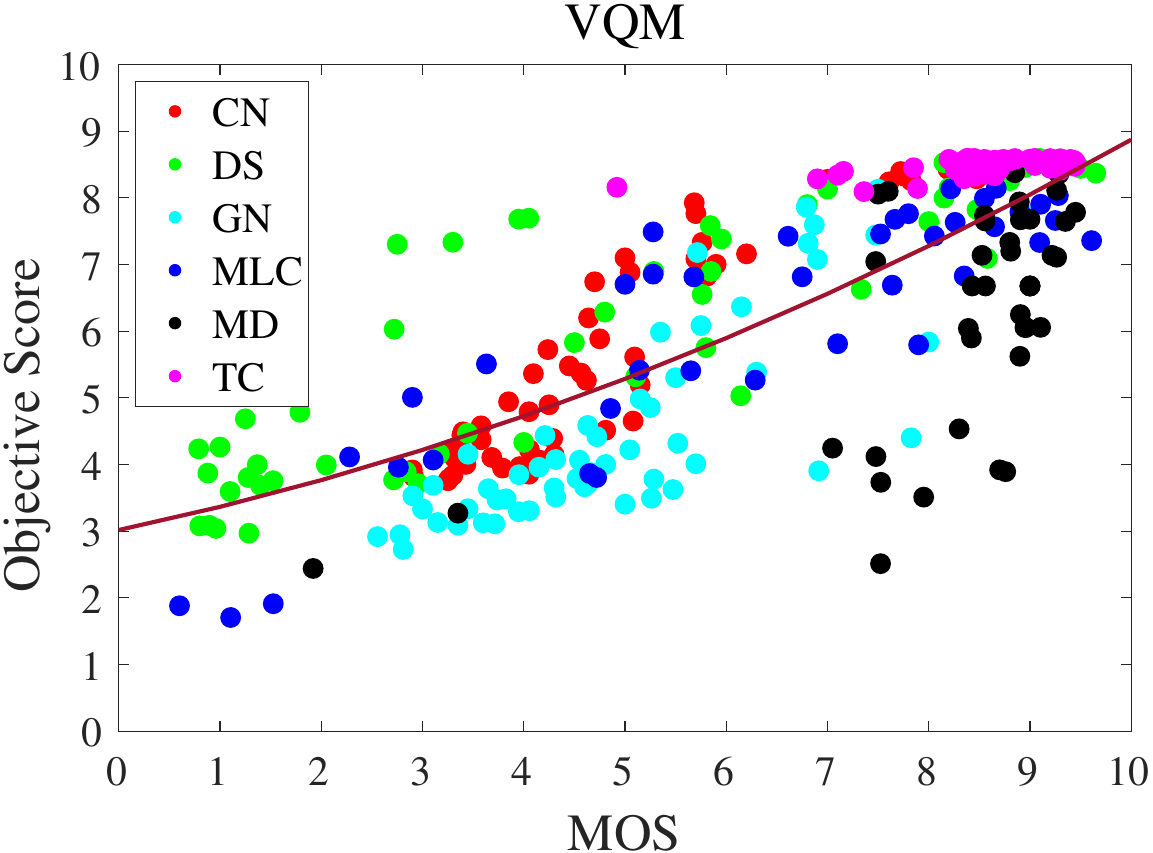}}	
	\caption{Scatter plots for the video-based metrics.}
\label{fig:video_based_curve}
\vspace{-0.2cm}
\end{figure}

Table \ref{Table:Metric-perfromance-1}-$\bf C$ presents the results of video-based metrics. We see that among video-based metrics, MS-SSIM reports the strongest overall results, with average PLCC and SROCC values of 0.90 and 0.88, similar to the performance of ${\rm PCQM}_{\rm p}$. SSIM and 3SSIM show close results, averaging around 0.85 and 0.80 for PLCC and SROCC, respectively. PSNR, VMAF, and VQM illustrate close performance, with overall PLCC and SROCC scores equal to or slightly below 0.8. To illustrate the reasons for the difference in performances among video-based metrics, we display the scatter plots of SSIM, MS-SSIM, VMAF, and VQM in Fig. ~\ref{fig:video_based_curve}, the red lines represent the best fitted curves. The objective scores are mapped to the 0-10 range using Eq. \eqref{eq:log}.  

After mapping the objective scores to the scale of MOS, an ideal scatter plot would have all its points distributed close to the ``y=x'' function (y representing objective scores, and x representing the MOS), indicating that the predicted objective scores are equal or very close to the subjective scores. Based on Fig. \ref{fig:video_based_curve}, we see that MS-SSIM shows the best distribution, explaining why it illustrates the best overall PLCC and SROCC. SSIM, VMAF, and VQM have partial scatters far away from the best fitted curves. For example, SSIM leans towards giving low scores for CN and high scores for DS, VMAF tends to score higher for DS and lower for MD, and VQM favors TC and DS while giving lower scores to MD. As shown in the analysis in section \ref{sec:mos-ana}, viewers tend not to be bothered by MD distortions and give them higher scores, which explains the lower performances of VMAF and VQM when compared to SSIM.

For the results of video-based metrics on different distortions, we see that all the video-based metrics present good performance for CN, DS, and GN. PSNR shows low correlations for MD, with an average PLCC of 0.58. SSIM reports poor performance for TC, having PLCC values of 0.48. MS-SSIM presents robust performance on CN, DS, GN, MLC, and MD distortions, the PLCC are higher than 0.90. VMAF also reveals good results on individual distortions, but as we analyzed in the previous paragraph, the relative values of VMAF scores for different distortions have some biases, e.g., lower scores for MD. VQM shows the worst PLCC on MLC (0.76) compared to other metrics. 3SSIM is the only metric that the PLCC of DS is lower than 0.90. However, it demonstrates good performance on MLC in which DS is introduced during texture map compression. Therefore, we think that 3SSIM can successfully detect other operations which can incur distortions such as Draco compression.

In summary, partial video-based metrics such as MS-SSIM show reliable  performance. However, some canonical metrics report opposite results to human perception for specific distortion (e.g., VMAF and VQM on MD).

\subsubsection{Summary of Objective Metrics}
${\rm PCQM}_{\rm p}$ and MS-SSIM report the best overall performance, with PLCCs and SROCCs of (0.91, 0.87) and (0.90, 0.88), respectively. MS-SSIM is more stable than PCQM among different types of distortion. For example, based on Table. \ref{Table:Metric-perfromance-1}, for PLCC, MS-SSIM has five results higher than 0.90 while ${\rm PCQM}_{\rm p}$ only has three, and the worst PLCC of MS-SSIM is 0.74 which is also higher than ${\rm PCQM}_{\rm p}$'s worst of 0.66. If these two metrics have close overall results, they also present advantages and drawbacks. The performance of ${\rm PCQM}_{\rm p}$ could be affected by the resolution of the sampling method used. For MS-SSIM, extra time is needed to render the DCM as PVSs. Moreover, the camera path used to render the PVSs is handcrafted and worth pondering. For different types of DCM, viewers may have different regions of interest. It is assumed that different camera paths could lead to different prediction results. However, when meshes share similar characteristics, as in the proposed test, applied to DCM, this phenomenon is less likely to happen, relying on MS-SSIM is a good option. Otherwise, we believe that ${\rm PCQM}_{\rm p}$ is a more appropriate choice.

\subsection{The Influence of Sampling on Point-based Metric} \label{sec: exp_sampling}

As outlined in the experimental results of section \ref{sec:point-based-performance}, point-based metrics (particularly ${\rm PCQM}_{\rm p}$) showcase high performance on DCM quality assessment. An advantage of these metrics over image-based and video-based metrics is that they do not need rendering: they extract features from raw mesh data, which generally have less computing time overhead. However, these metrics rely on  mesh sampling, and little research has been conducted on the influence of sampling on DCM quality assessment \cite{fu2023surface}. Therefore, in this section, we first present the performance of point-based metrics with various grid sampling resolutions, then compare the differences in performances between sampling methods.

\subsubsection{Grid Sampling Resolution} 
\begin{figure}[h]
    \centering
    \includegraphics[width=1\linewidth]{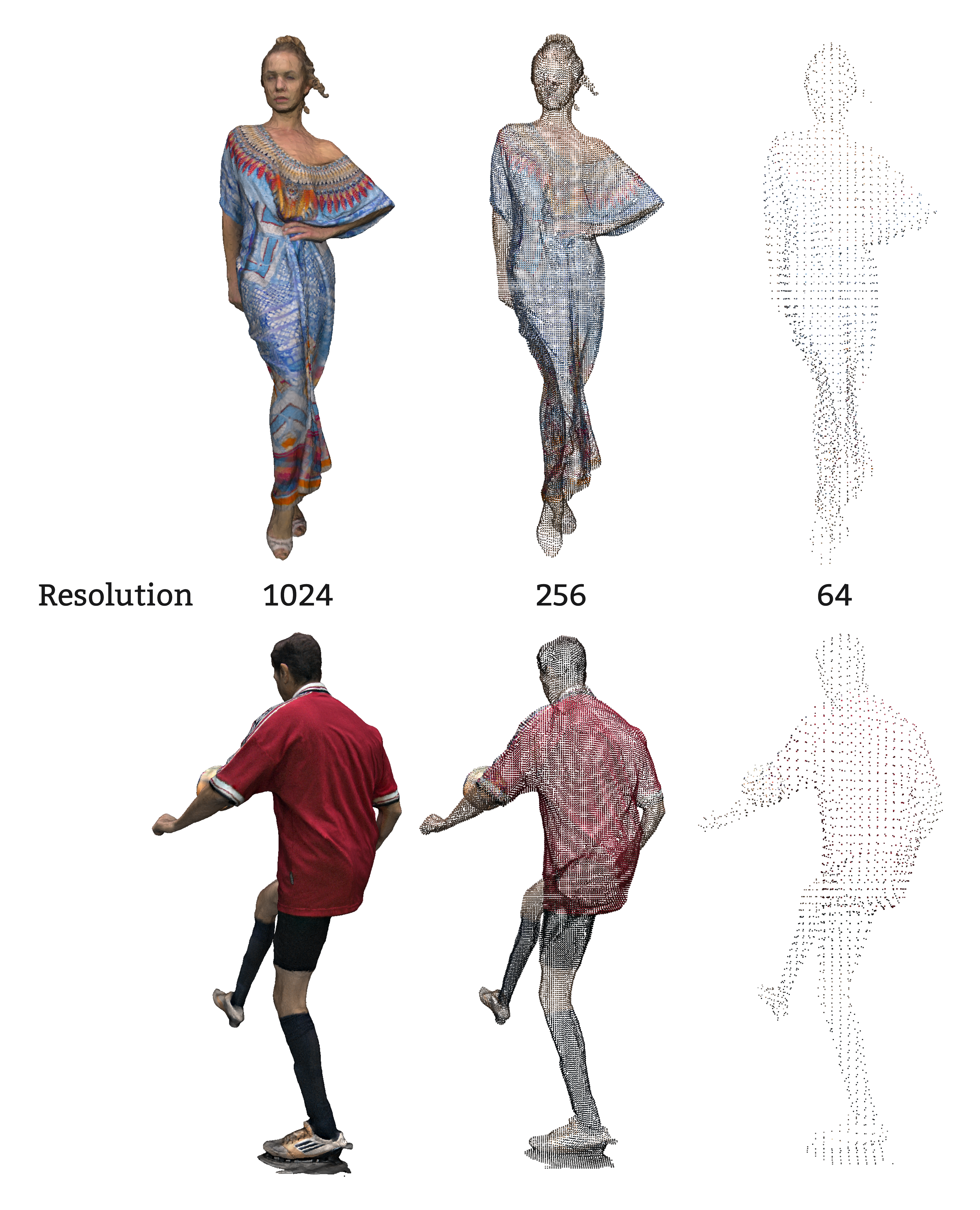}
    \caption{Grid sampling results concerning different resolutions.}
    \label{fig:sampling}
\end{figure}

The goal of this part is to study the correlation between grid sampling resolution and metric accuracy. We propose eight different grid resolutions: 1024, 512, 256, 128, 64, 32, 16, and 8. Table \ref{tab:grid-sampling} illustrates the sampled point clouds of grid sampling with different resolutions, which will be used for the calculation of point-based metrics. We use the first frame of ``Longdress'' and ``Football'' as examples to reveal the variation of the number of sampled points. Fig. \ref{fig:sampling} presents the visual impact of using fewer points. We can observe that with the decrease of sampling resolutions, the point clouds become sparser.

    \begin{table}[h]
	\caption{Examples of sampled point clouds with different grid resolutions} \label{tab:grid-sampling}
	\centering
 \begin{scriptsize}
 \renewcommand{\arraystretch}{1.5}
	\setlength{\tabcolsep}{0.5mm}{
	\begin{tabular}{|c|c|c|}
		\hline
		 \multirow{2}{*}{Resolution}&\multicolumn{2}{c|}{Points}\\\cline{2-3}
            & Longdress & Football  \\ \cline{1-3}
		1024 &577,679&	548,633  \\ \cline{1-3}
		512 & 144,218&	136,920  \\ \cline{1-3}
		256 & 35,912&	34,126  \\ \cline{1-3}
            128 & 8,974&	8,501 \\ \cline{1-3}
            64 & 2,189&	2,045 \\ \cline{1-3}
            32& 528&	497 \\ \cline{1-3}
            16& 120&	113 \\ \cline{1-3}
            8 & 22	&17 \\ \cline{1-3}
	\end{tabular}}
 \end{scriptsize}
\end{table}

\begin{figure}[t]
	\centering
	\vspace{-0.35cm}
	\subfigtopskip=2pt
	\subfigbottomskip=2pt
	\subfigcapskip=-5pt
	\subfigure{
		\includegraphics[width=0.45\linewidth]{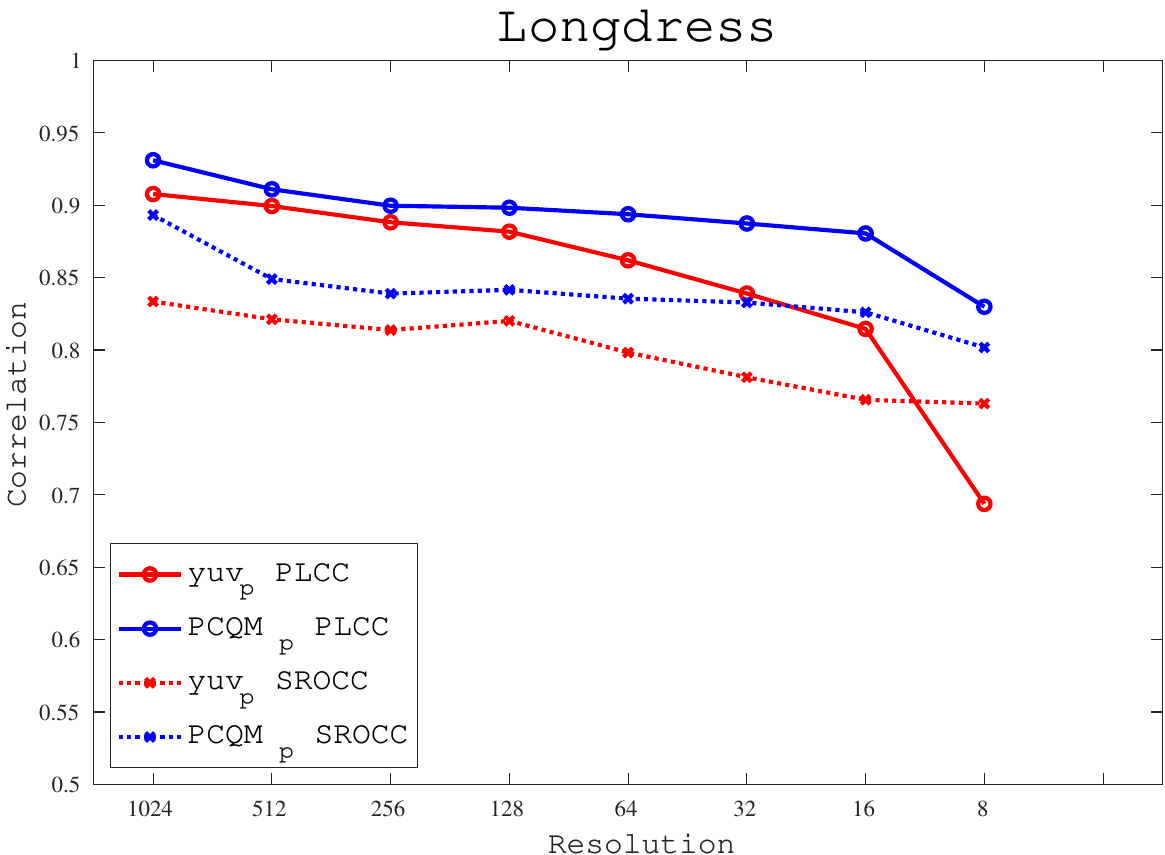}}	
	\subfigure{		
		\includegraphics[width=0.45\linewidth]{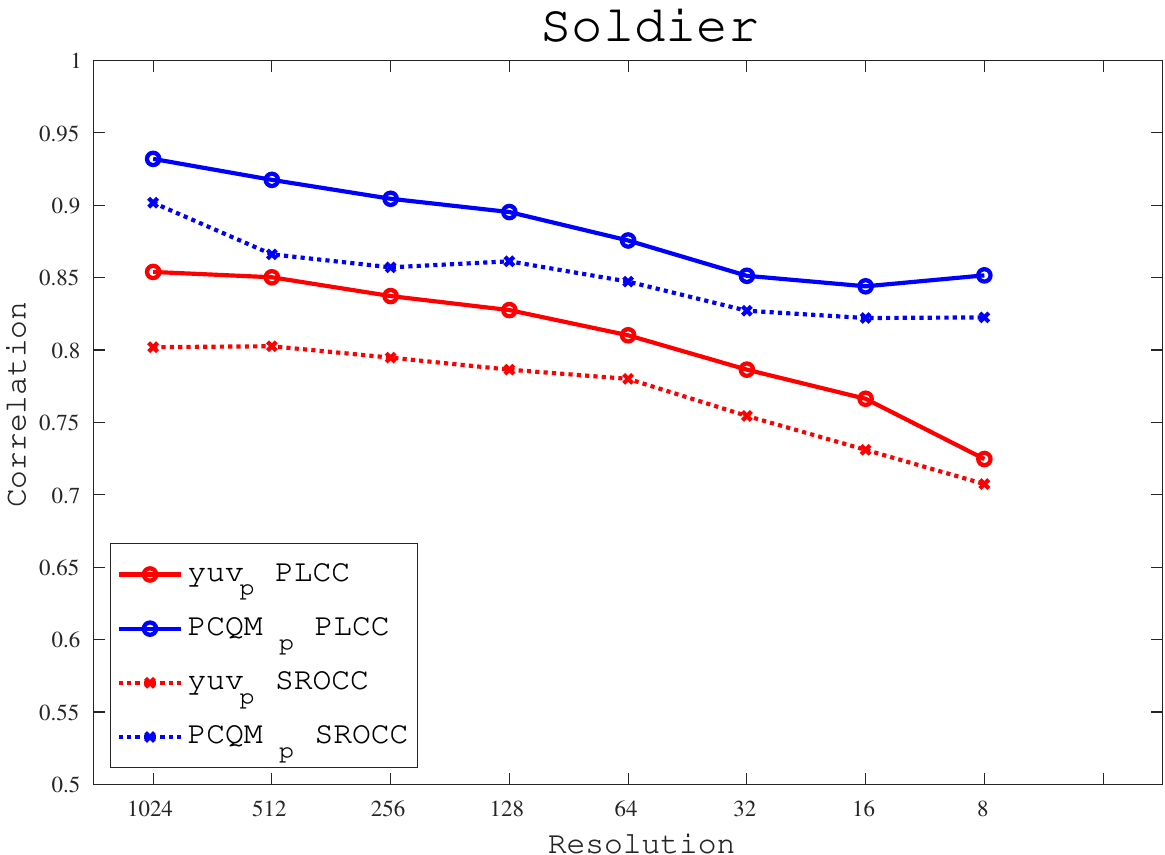}}   
	\subfigure{	
		\includegraphics[width=0.45\linewidth]{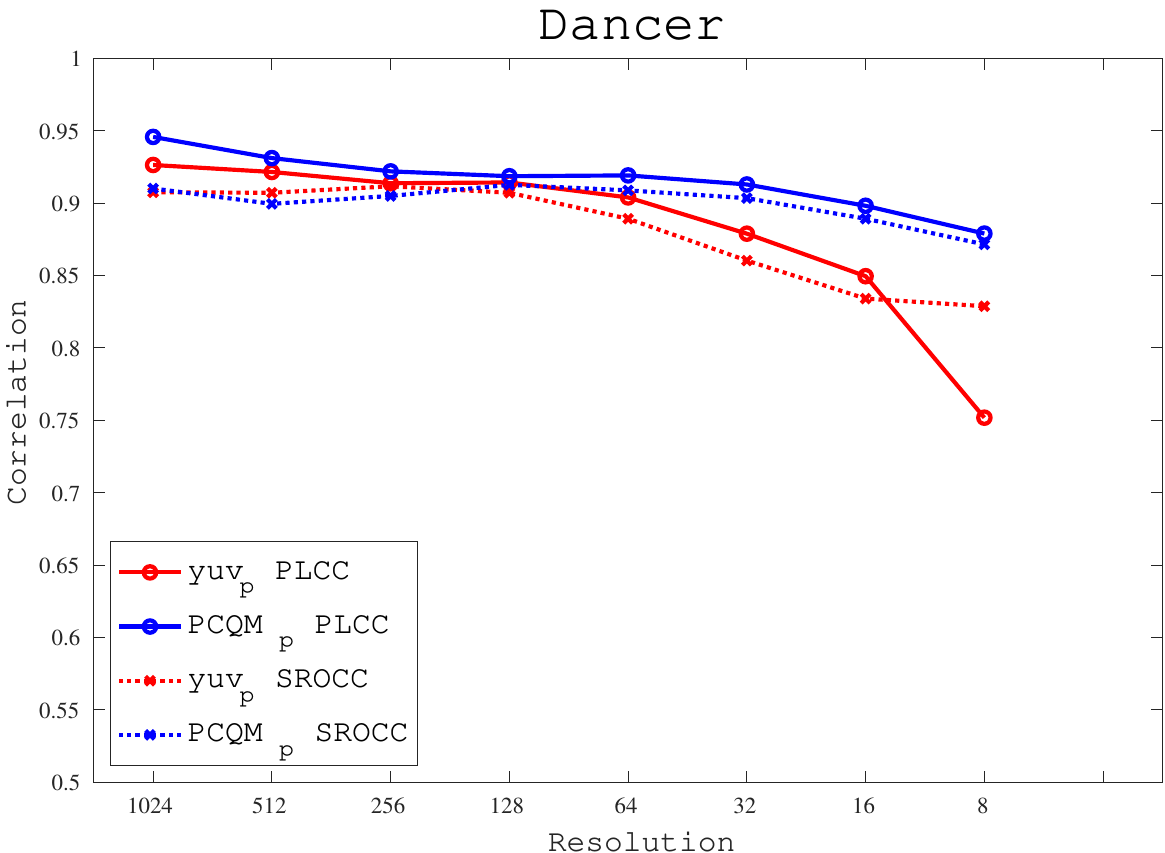}}	
	\subfigure{		
		\includegraphics[width=0.45\linewidth]{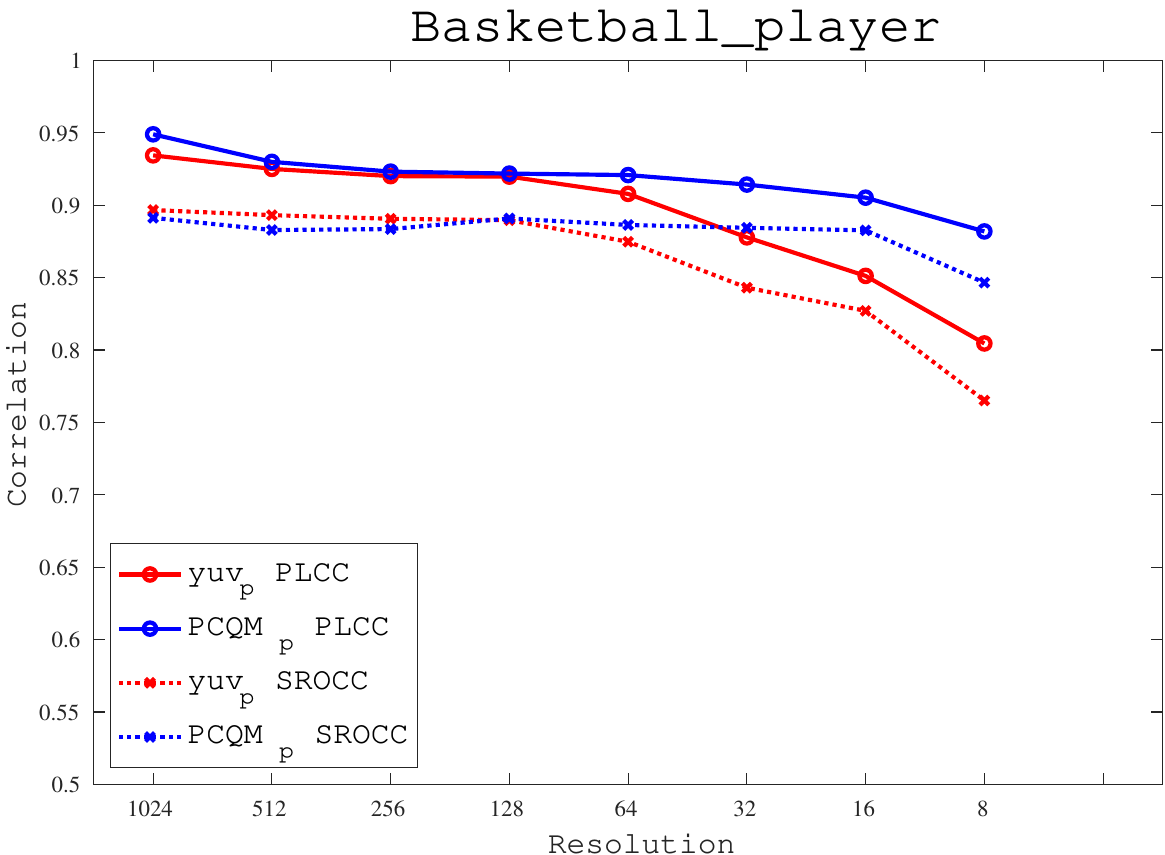}}	
  	\subfigure{
		\includegraphics[width=0.45\linewidth]{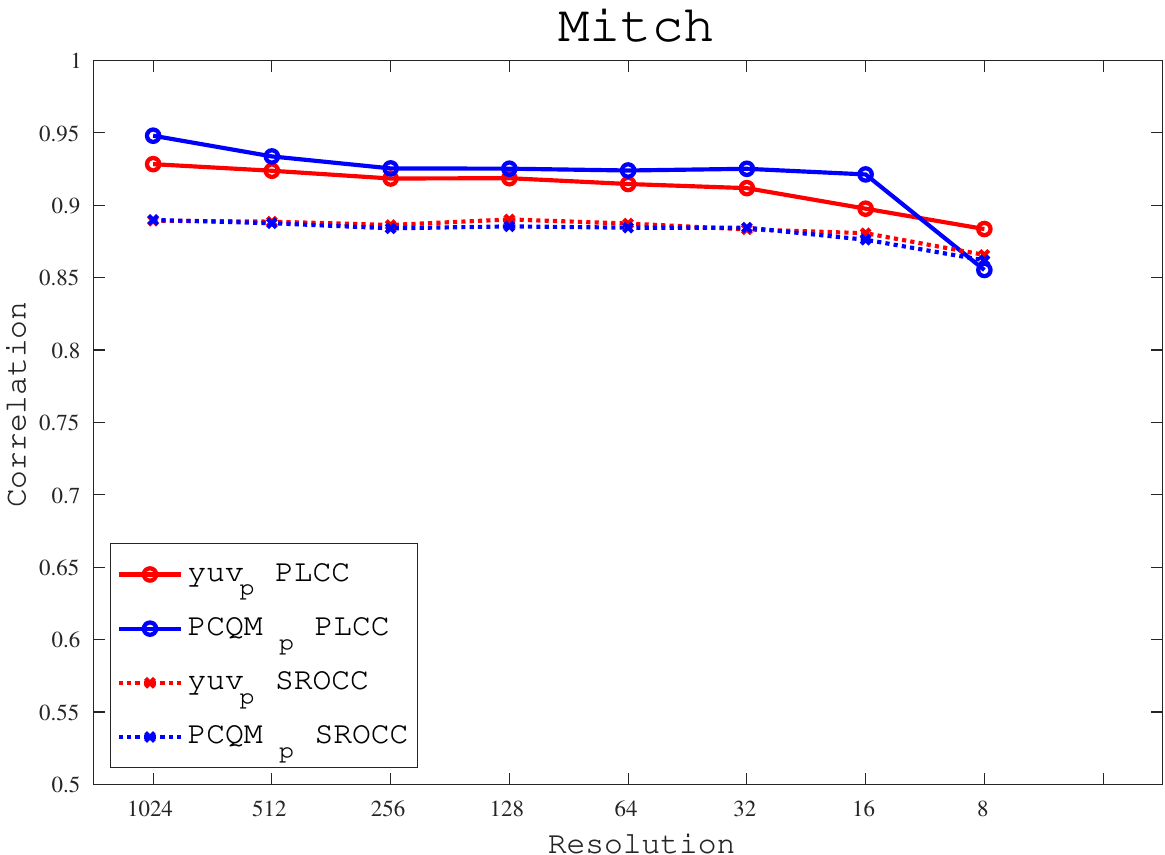}}	
	\subfigure{		
		\includegraphics[width=0.45\linewidth]{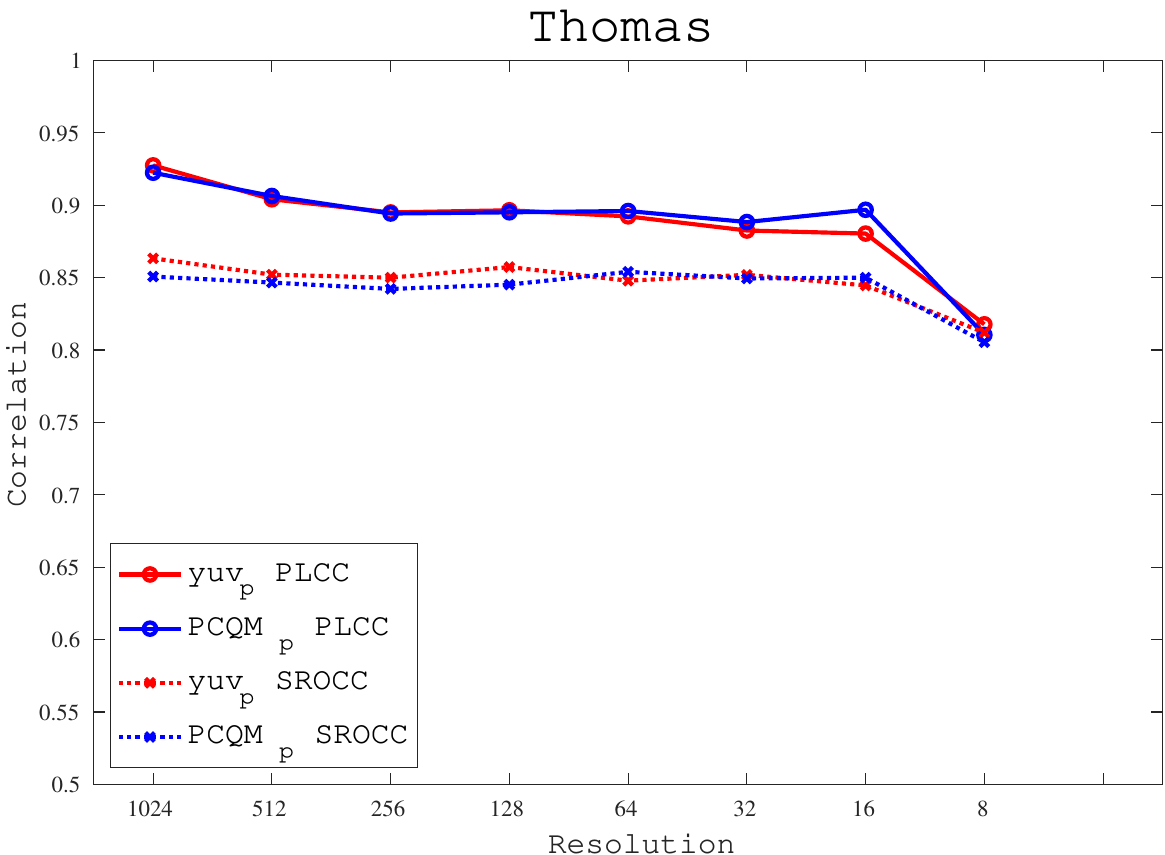}}   
	\subfigure{	
		\includegraphics[width=0.45\linewidth]{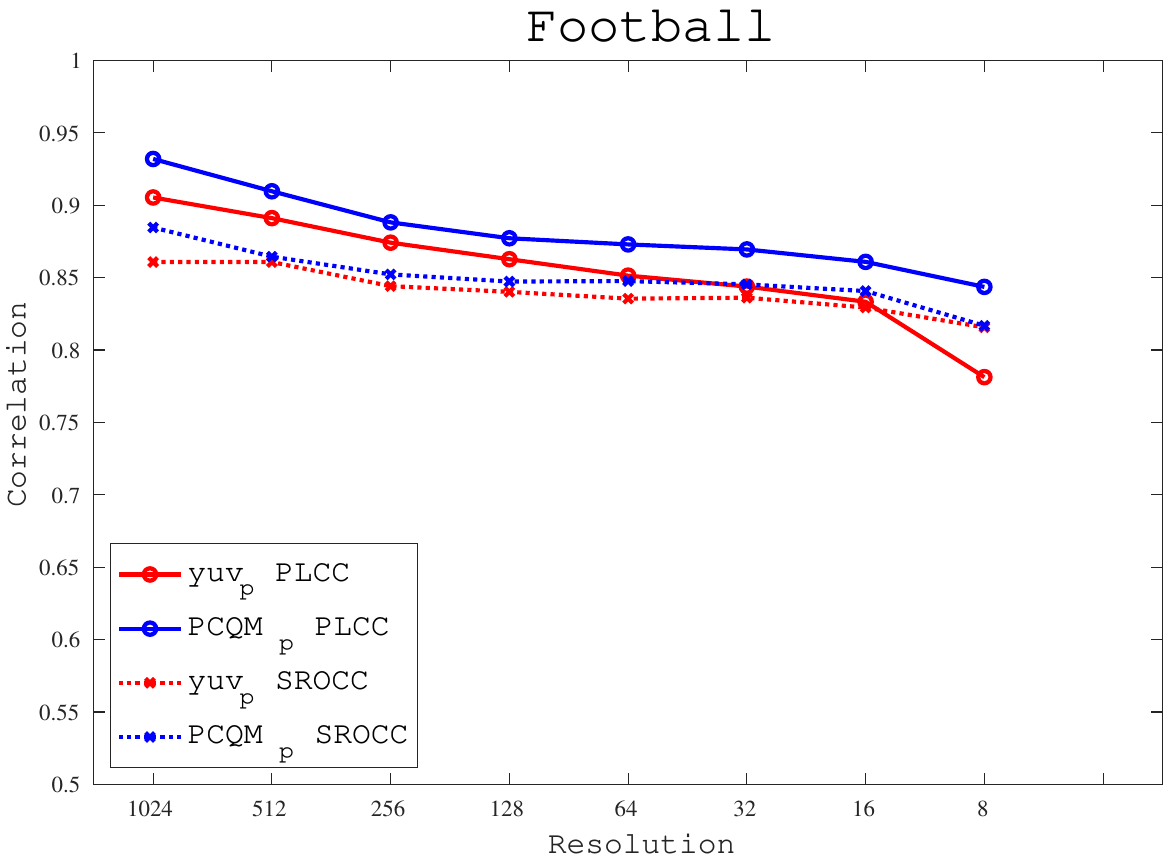}}	
	\subfigure{		
		\includegraphics[width=0.45\linewidth]{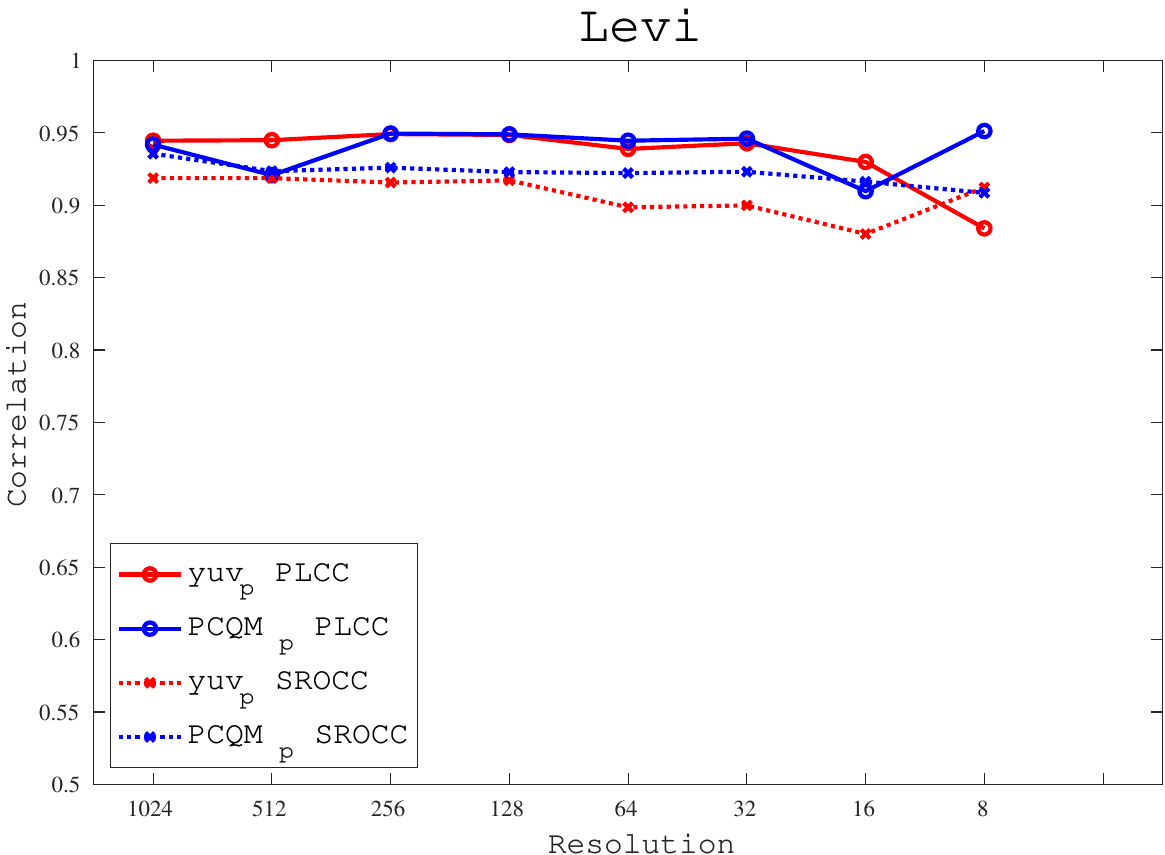}}
	\caption{Performance of $\rm yuv_{p}$ and ${\rm PCQM}_{\rm p}$ with different grid sampling resolutions on different sequences.}
	\label{fig:metric-sampling-performance}
\end{figure}

We plot the performance of $\rm yuv_{p}$ and ${\rm PCQM}_{\rm p}$ with regards to several grid sampling resolutions on eight sequences in Fig. \ref{fig:metric-sampling-performance}. First, we see that both $\rm yuv_{p}$ and ${\rm PCQM}_{\rm p}$ tend to show worse performance with the decrease of sampling resolution. One exception is ``Levi'', the PLCC of $\rm yuv_{p}$ and the SROCC of $\rm yuv_{p}$ are increased when resolution decreases from 16 to 8. We think the reason is that: when sampling resolutions blow 32, the sampled point clouds are too sparse to calculate effective structure feature. For example, ${\rm PCQM}_{\rm p}$ uses a radius search to construct a local patch to calculate the local curvature. If the patch only has a few or one points, the calculated feature might not represent local curvature anymore. Second, compared to other sequences, grid sampling obviously showcases stable performance on ``Levi''. We hypothesize that the main reason is that the texture of ``Levi'' is relatively simple compared with other sequences, e.g., the black sweater, the black shoes, and the dark skin, making it an easier case for sampling.

\begin{figure*}[b]
    \centering
    \includegraphics[width=1\linewidth]{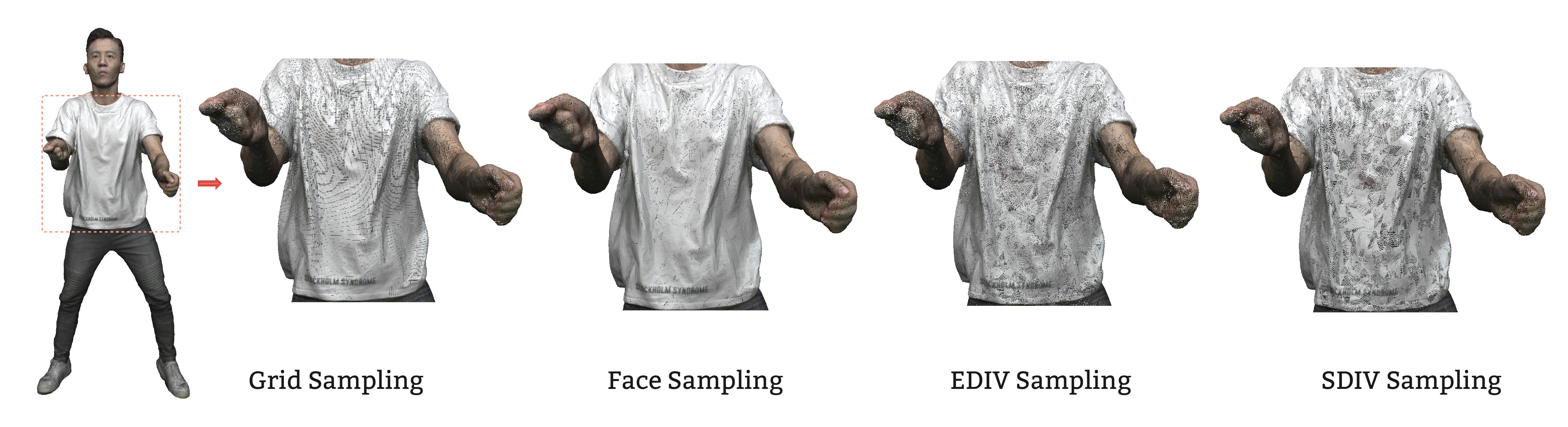}
    \caption{Snapshots of different sampling methods.}
    \label{fig:different-sampling-method}
\end{figure*}
Generally, the denser the sampled point clouds, the closer the perception to the original DCM.  In practical applications, to better balance accuracy and computational complexity for point-based metrics, we recommend using grid sampling with a resolution between 1024 and 256 for DCM samples sharing similar characteristics as TDMD samples.

\subsubsection{Different Sampling Methods}

We study the results of point-based metrics on four different sampling methods proposed in \cite{MPEG-MESH-metric}: grid sampling, face sampling, surface subdivision sampling (SDIV), and edge subdivision sampling (EDIV).  Grid sampling generates the points based on the intersections between faces and a 3D grid, face sampling generates the points face by face using a 2D grid following the direction of the triangle edges (the directions of the edges of 2D grid are oriented parallel to the two edges of the triangular face, respectively). SDIV recursively subdivides faces until an area threshold is reached, while EDIV recursively subdivides faces until an edge length threshold is reached. 
The detailed introduction of these sampling methods can be found in \cite{MPEG-MESH-metric}.

We show an overview of the sampled point clouds corresponding to the four sampling methods in Fig. \ref{fig:different-sampling-method}. To better illustrate the characteristics of these methods, we zoom in on the contents of the orange-dotted rectangle box. We see that grid sampling incurs horizontal and vertical cracks. The reason for this phenomenon is that the generation of a point relies on whether the face intersects the 3D grid, and thus the density of the points is affected by the grid resolution. If a face is close to or smaller than the smallest unit of the grid, few points are sampled, which causes strip cracks. Face sampling exhibits  some triangular areas with ``uneven'' point density. Face sampling samples points face by face using the same resolution, which means that the point density is identical for each triangle patch.  The angle between the viewing direction and the normal vector direction of each face is different, causing the perceived point density to be different. EDIV and SDIV present irregular cracks because they also sample points face by face but use a threshold to indicate the triangle edge and the size subdivision endpoint, respectively. Considering that the face edges and sizes are different, the sampling results illustrate different and irregular point distributions.

After previously reporting the performances of grid sampling, we will now study the results of face sampling, SDIV, and EDIV. These three sampling methods all sample points face by face. For face sampling, the minimum number of sampled points is equal to 1/3 of the number of mesh vertices; for SDIV and EDIV, the minimum number is equal to the number of mesh vertices. We set seven sampling grades for each sampling method: for face sampling and EDIV, we set sampling resolution to 1024, 512, 256, 128, 64, 32 and 16; for SDIV, we use the 1/16, 1/8, 1/4, 1/2, 1, twofold, and fourfold average face size of each sequence as the thresholds. For the same grade of different sampling methods, the sampled point clouds might have a different number of points. For example, for the first grade, using the first frame of ``Longdress'' as an example, face sampling, EDIV, and SDIV have 1200K, 1260K, and 1460K points, respectively.

\begin{table*}[t]
	\caption{Metrics performance (PLCC and SROCC ) for the proposed database with different sampling methods. } \label{Table:different sampling}
	\centering
	\begin{scriptsize}
  \renewcommand{\arraystretch}{1.5}
	\setlength{\tabcolsep}{0.7mm}{
		\begin{tabular}{|c|c|c|c|c|c|c|c|c|c|c|c|c|c|c|c|c|c|c|}
			\hline
   Grade &\multicolumn{6}{c|}{I}&\multicolumn{6}{c|}{II}&\multicolumn{6}{c|}{III}\\ \hline
			{Index}& \multicolumn{3}{c|}{PLCC} & \multicolumn{3}{c|}{SROCC}& \multicolumn{3}{c|}{PLCC} & \multicolumn{3}{c|}{SROCC}& \multicolumn{3}{c|}{PLCC} & \multicolumn{3}{c|}{SROCC}  \\ \hline
			{Metric}&Face&EDIV&SDIV&Face&EDIV&SDIV&Face&EDIV&SDIV&Face&EDIV&SDIV&Face&EDIV&SDIV&Face&EDIV&SDIV\\  \hline		





$\rm yuv_{p}$&0.82 &	0.82 	&0.81 &	0.80 	&0.79 &	0.79 &0.81 &	0.80 &	0.79 &	0.79& 	0.77 &	0.77 &	0.78 &	0.71 &	0.70& 	0.76& 	0.69 &	0.68 	 	 \\ \hline

${\rm PCQM}_{\rm p}$&0.91 &	0.91 &	0.91 	&0.88 &	0.87 &	0.87 &0.90 &	0.88 &	0.88 &	0.86 &	0.83 &	0.83 &0.86 &	0.77 &	0.77& 	0.82 &	0.74 &	0.74 		 	 \\ \hline
	\end{tabular}}
	\end{scriptsize}

\end{table*} 

\begin{figure}[t]
	\centering
	\vspace{-0.35cm}
	\subfigtopskip=2pt
	\subfigbottomskip=2pt
	\subfigcapskip=-5pt
	\subfigure{
		\includegraphics[width=0.45\linewidth]{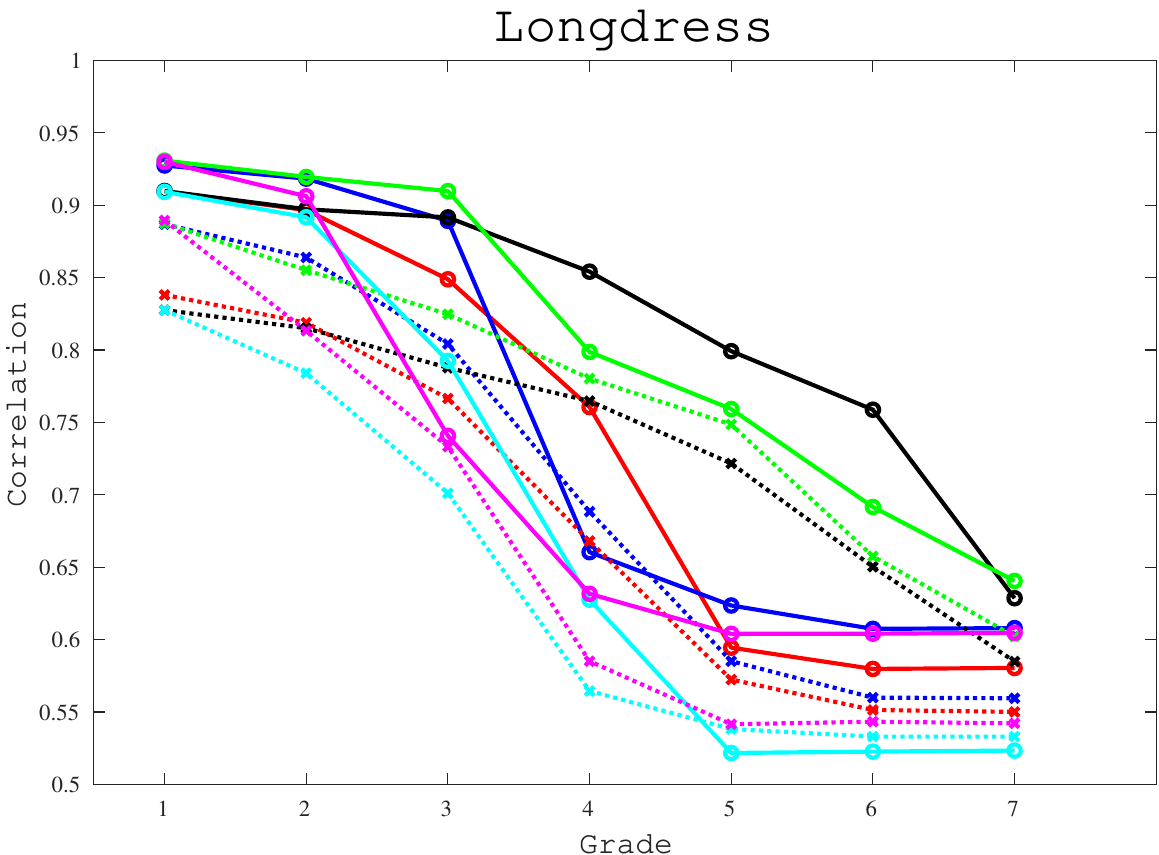}}	
	\subfigure{		
		\includegraphics[width=0.45\linewidth]{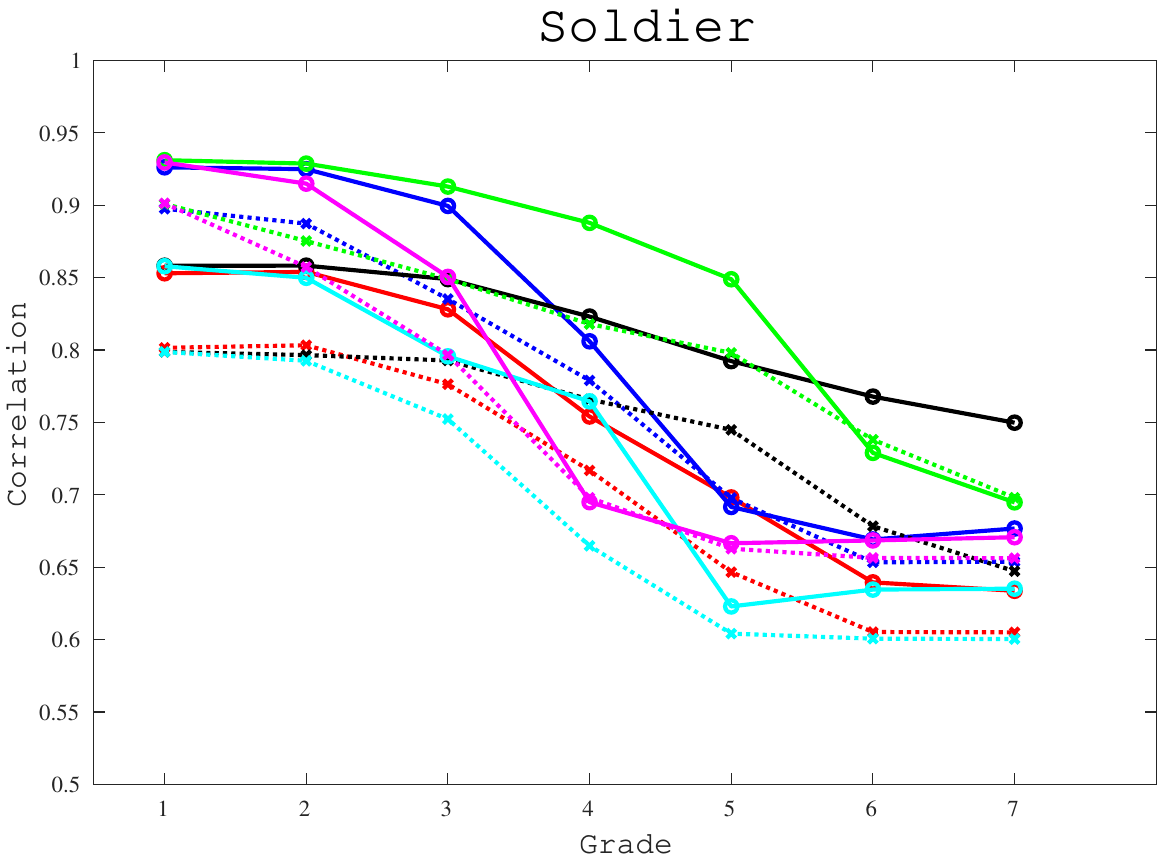}}   
	\subfigure{	
		\includegraphics[width=0.45\linewidth]{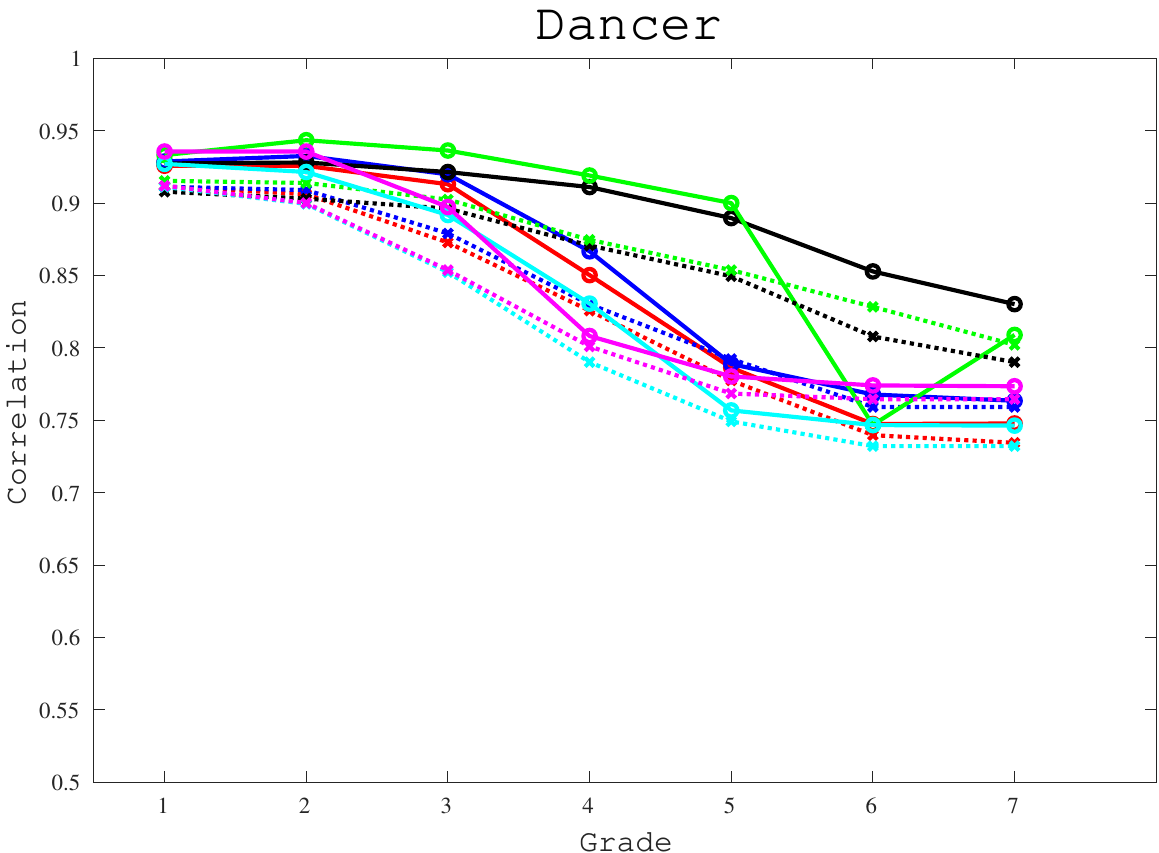}}	
	\subfigure{		
		\includegraphics[width=0.45\linewidth]{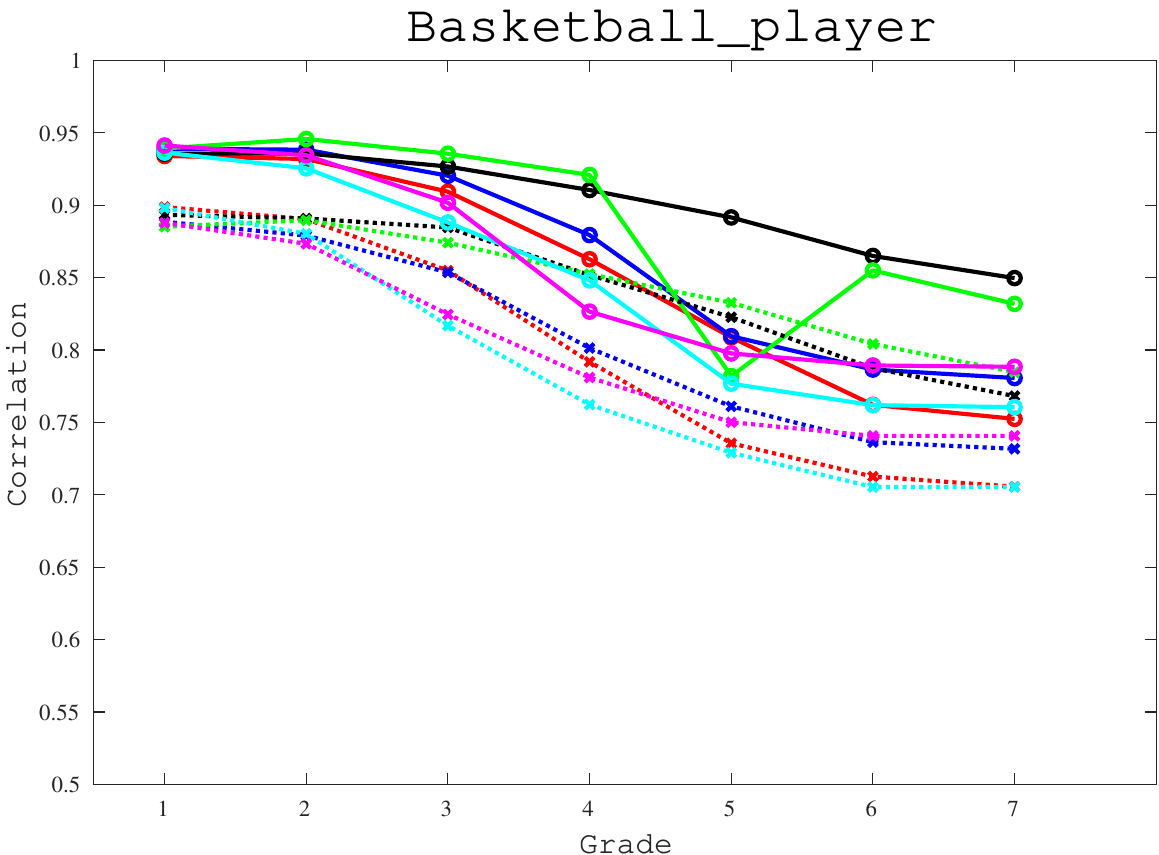}}	
  	\subfigure{
		\includegraphics[width=0.45\linewidth]{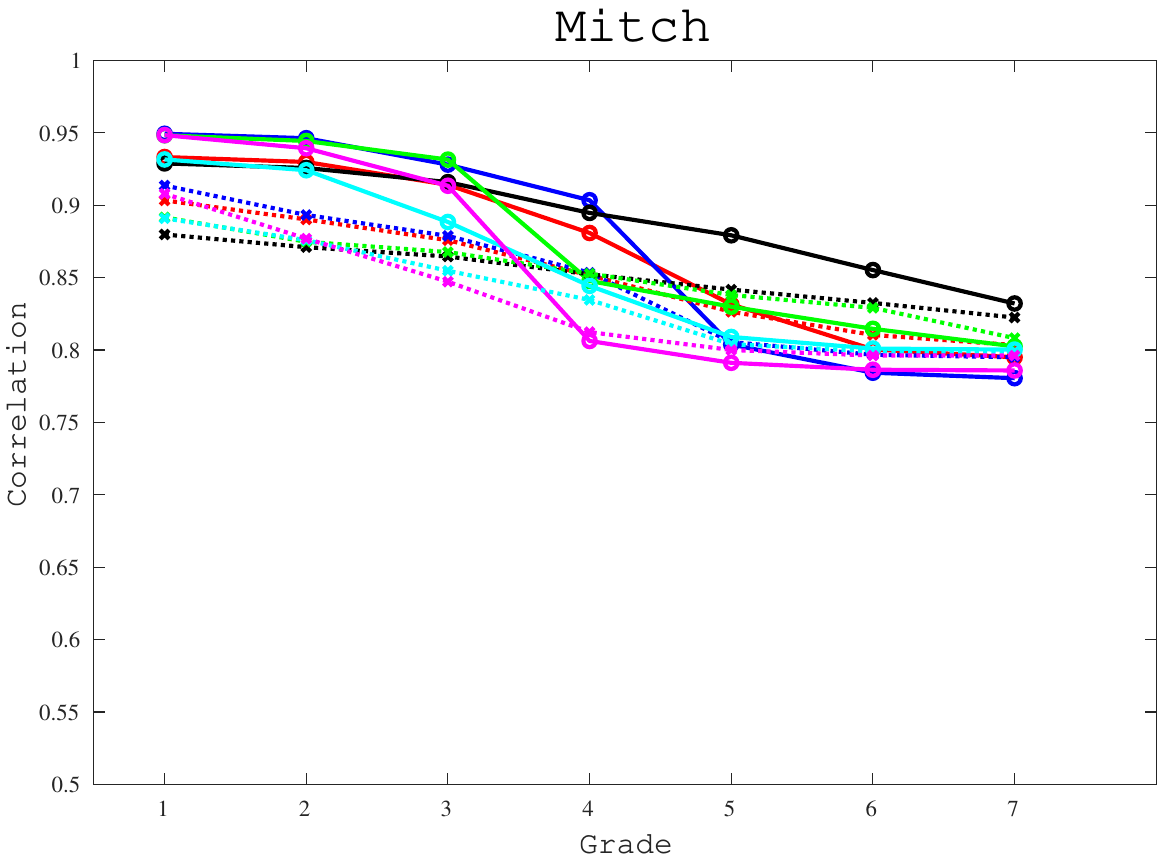}}	
	\subfigure{		
		\includegraphics[width=0.45\linewidth]{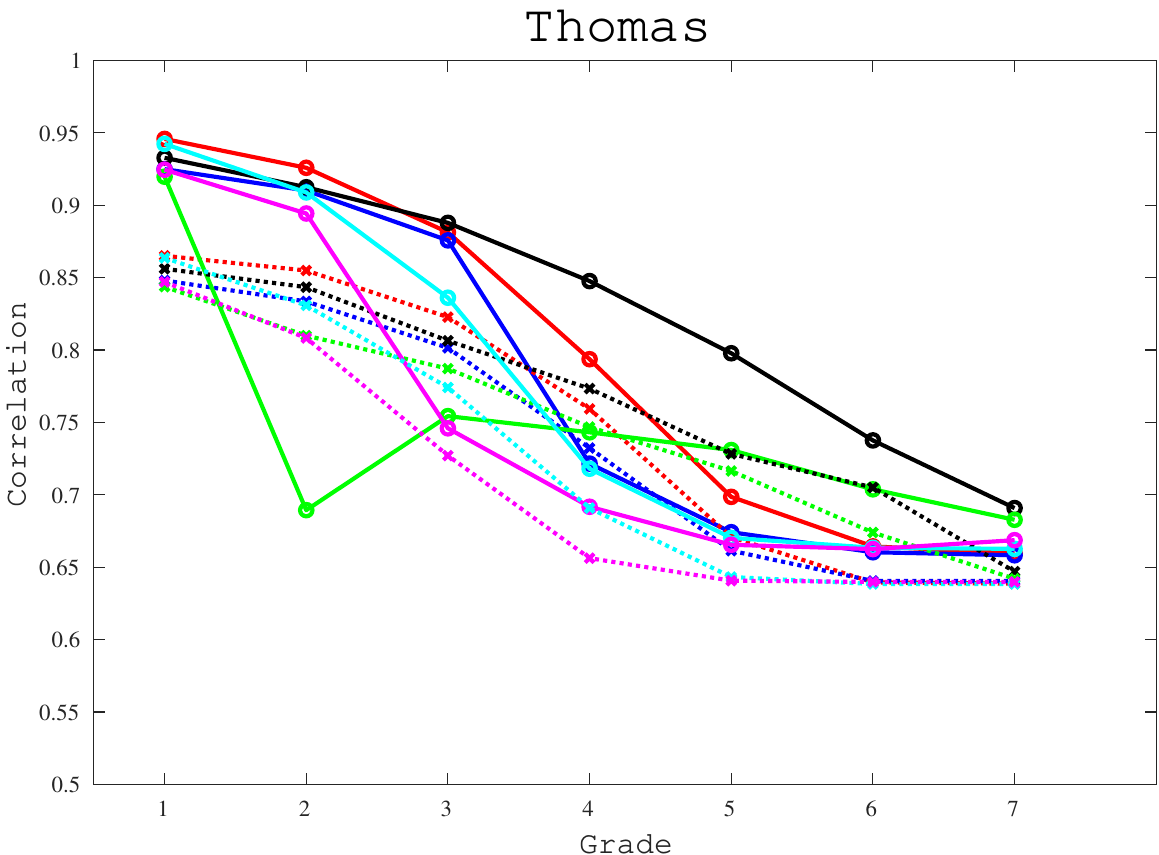}}   
	\subfigure{	
		\includegraphics[width=0.45\linewidth]{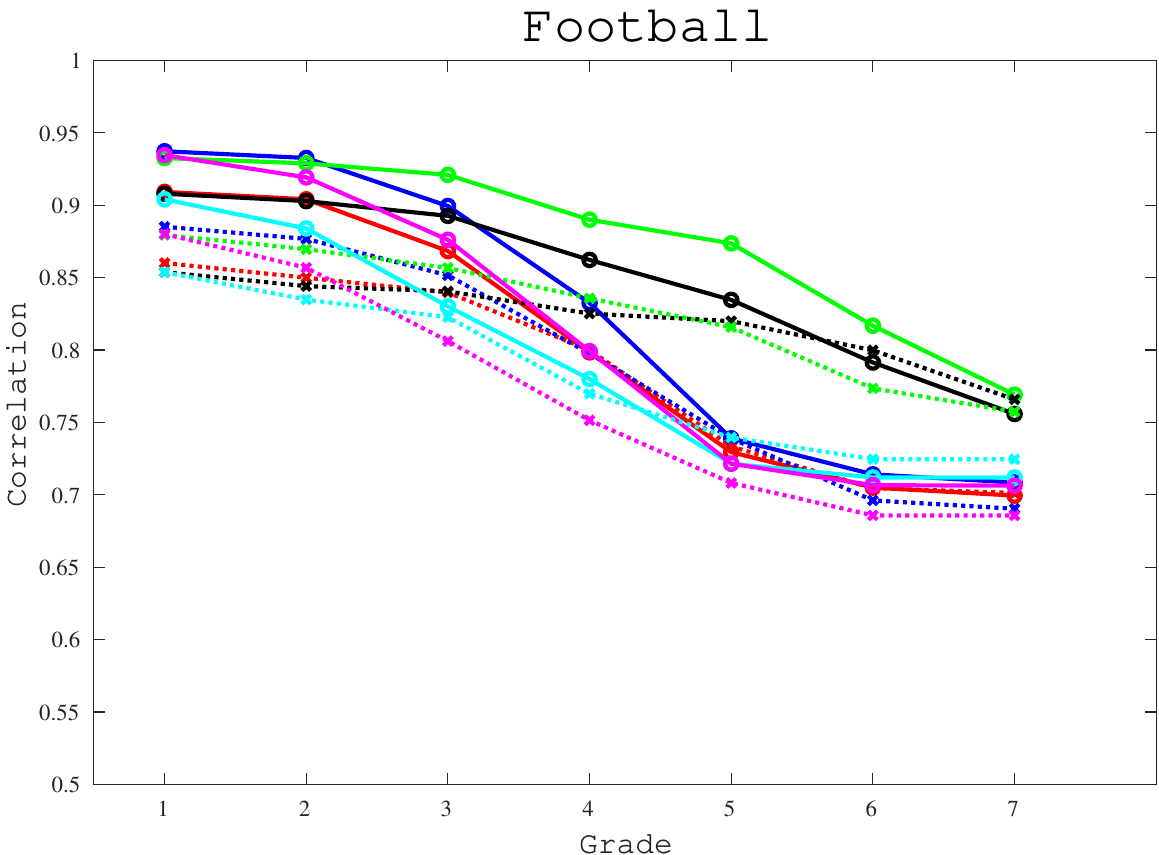}}	
	\subfigure{		
		\includegraphics[width=0.45\linewidth]{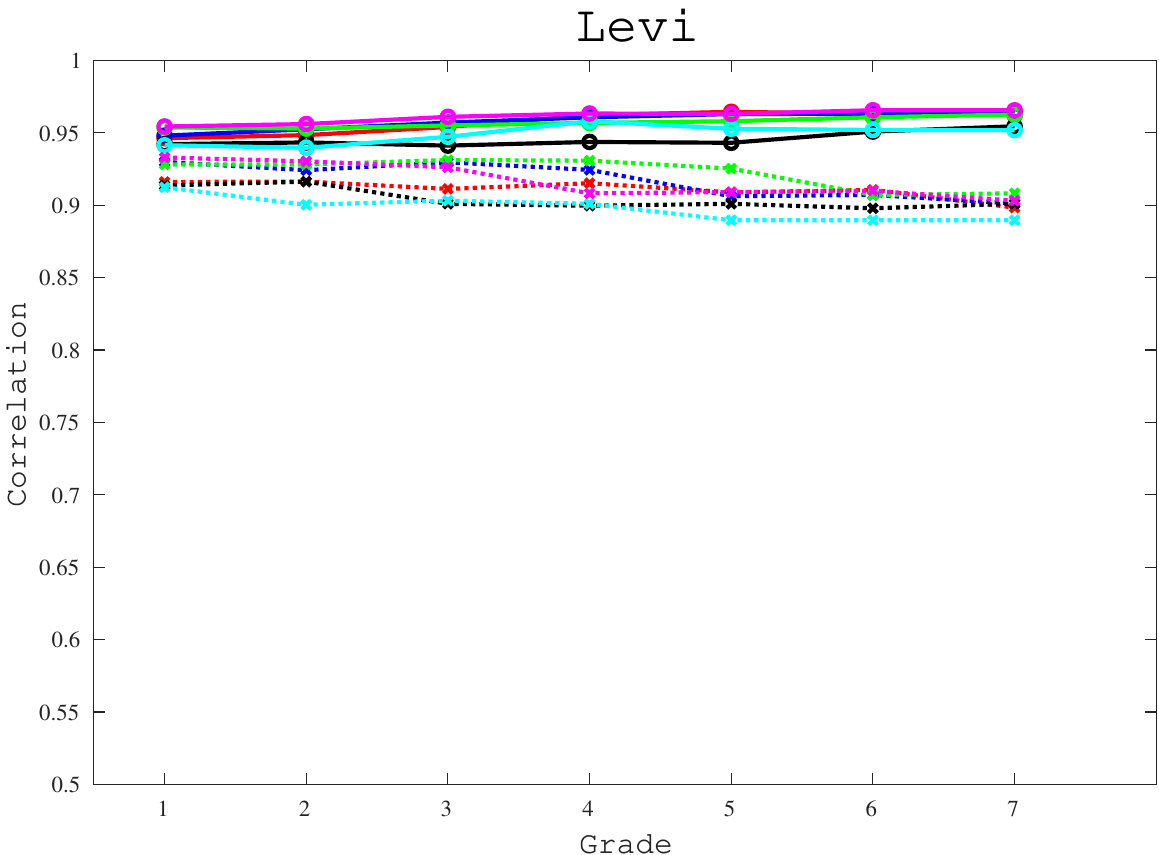}}
  	\subfigure{		
		\includegraphics[width=0.8\linewidth]{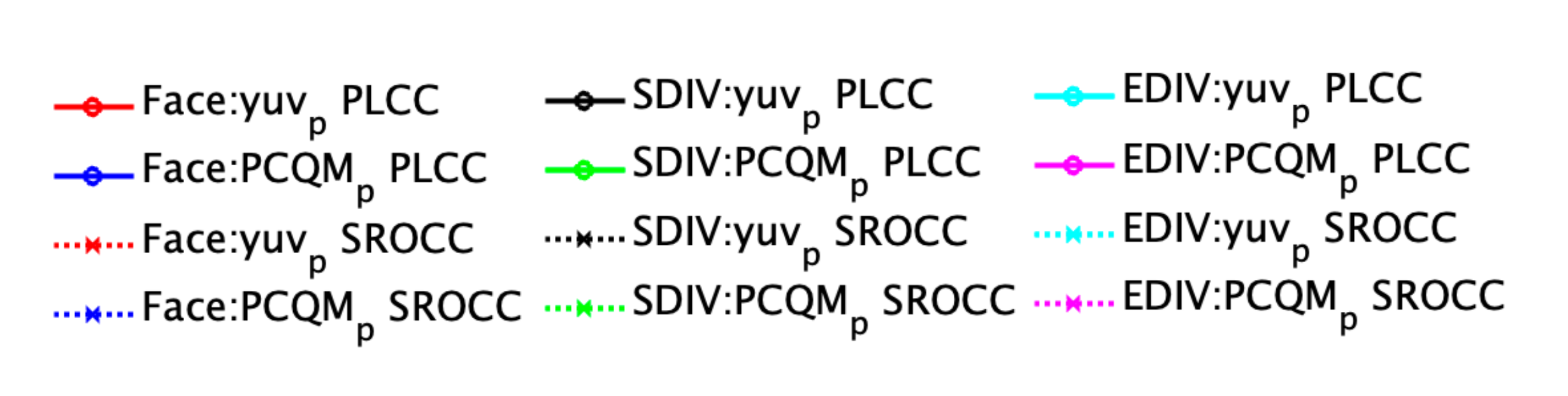}}
  
	\caption{Performance of $\rm yuv_{p}$ and ${\rm PCQM}_{\rm p}$ with different sampling methods on different sequences.}
	\label{fig:metric-sampling-method-performance}
\end{figure}

The results are shown in Fig. \ref{fig:metric-sampling-method-performance}. We see that with the decrease of the number of points (that is, the increase in grade), $\rm yuv_{p}$ and ${\rm PCQM}_{\rm p}$ tend to show worse results regardless of the types of sampling methods. The results of ${\rm PCQM}_{\rm p}$ with SDIV presents more unstable performance than face sampling and EDIV for sequence ``Dancer'', ``Basketball\_player'' and ``Thomas'' (cf., the green line). Three sampling methods might present unchanged results for the last several grades on some sequences, suggesting that the increase of sampling grade will not change the sampling results anymore, i.e., the sampled point clouds have reached the minimum points. ``Levi'' again illustrates more stable results than other sequences with regards to different sampling methods, which shares the same reason as grid sampling.

The performance of point-based metrics might be influenced by the point density. To ensure a relatively fair performance comparison, the sampled point clouds between different sampling methods should have a close number of points. 
We choose three sampling grades (I, II, and III) representing a number of points of around 1200K, 400K, and 100K, respectively. For face sampling, the corresponding sampling resolutions are 1024, 512, and 256. For EDIV,  the corresponding sampling resolutions are 1024, 512, and 128. For SDIV, the thresholds are 1/16, 1/4, and fourfold average face size of each sequence.

The results are shown in Table \ref{Table:different sampling}. Three sampling methods reveal close performance for grade I and II.  Therefore, if the sampled point clouds are dense enough, the types of sampling methods will have a reduced effect on the point-based metrics. Face sampling presents better performances than EDIV and SDIV in grade III, which is a relatively sparse sampling result. The sparser the point clouds, the less computationally complex the point-based metrics are. Taking into account all the results from Table \ref{Table:different sampling}, we recommend using face sampling for low-latency applications.

\section{Conclusion}\label{sec:conclusion}
In this paper, we first propose a new DCM database which contains 303 samples, called TDMD. It gathers eight reference sequences and six types of distortions: color noise, texture map downsampling, geometrical Gaussian noise, mesh decimation, MPEG lossy compression, and texture map compression. We have used a predefined camera path to convert DCM to PVS and have conducted extensive subjective experiments. We analyze the influence of different distortions on DCM perception. The MOSs of CN, DS, MLC, and GN present wide variations, while MD and TC seem to have limited impact on visual perception. We think the levels of MD suggested by WG7, and considered in this paper are too low to cause obvious perception degradation, and fine texture distortion caused by TC is masked due to the frame switching of dynamic sequence. 

Then, we test three types of objective metrics on the database, including image-based, point-based and video-based metrics.  We present the advantages and drawbacks of different metrics. ${\rm PCQM}_{\rm p}$ and MS-SSIM report the best overall performance. Image-based and video-based metrics need rendering before calculation which is time-consuming, while point-based metrics rely on the results of sampling which might reduce the overall performance. Based on the above results, we give suggestions about the selection of the metric in practical applications. 


There are many interesting tracks for future study. Besides viewing DCM via a 2D monitor, rendering DCM via a VR headset in an immersive environment is another option that can illustrate the different characteristics of DCM as a 3D media. Whether the viewers will report different subjective perceptions between the 2D monitor and the VR environment, and whether the objective metrics will show consistent results for these two viewing conditions deserve further exploration.

\appendix  

\subsection*{MLC parameters}
Table \ref{tab:target-bitrates} is the target bitrates of MPEG WG7 all-intra lossy compression mode, and the Draco quantization parameters for the position (Draco-QP) and texture coordinates (Draco-QT) vary from one sample to another.
    \begin{table}[h]   
    \small
	\caption{Target bitrates for all-intra lossy compression mode.} \label{tab:target-bitrates}
	\centering
	\begin{tabular}{|c|c|c|c|c|c|}
		\hline
		Test Sequence & R1 & R2 & R3 &R4 &R5 \\ \cline{1-6}
		Longdress & 5 & 9 & 12 &15 &22 \\ \cline{1-6}
		Soldier & 5 & 9 & 12 &15 &22 \\ \cline{1-6}
		Basketball\_player & 3 & 5 & 10 &14 &21 \\ \cline{1-6}
            Dancer & 3 & 5 & 10 &14 &21 \\ \cline{1-6}
            Mitch &4 & 6 &11 & 16 &24  \\ \cline{1-6}
            Thomas & 4 & 6 &11 & 16 &24  \\ \cline{1-6}
            Football & 4 & 8 &12 & 17 &25  \\ \cline{1-6}
            Levi & $4^{*}$ & 8 &12 & 17 &25  \\ \cline{1-6}
	\end{tabular}
\end{table}

 For ``Levi'', R1 cannot be reached due to the decimation problem explained above. As seen in Table \ref{tab:target-bitrates}, R1 is the lowest bitrate and it requires the application of MD. Therefore, there are only four available target bitrates for Levi.
 
\subsection*{Analysis of MD on Human Perception}

We postulate that the main reason for the increase of MD degree has a limited impact on human perception are: { the MD degrees proposed by WG7 are low, meaning that the number of faces after MD is still large enough to preserve most sample characteristics; and for these MD samples, } contour flicker is the main manifestation in perception quality degradation, which is less distracting than other types of distortion.  

\begin{figure}[h]
\setlength{\abovecaptionskip}{0.cm}
\setlength{\belowcaptionskip}{-0.cm}
    \centering
    \includegraphics[width=1\linewidth]{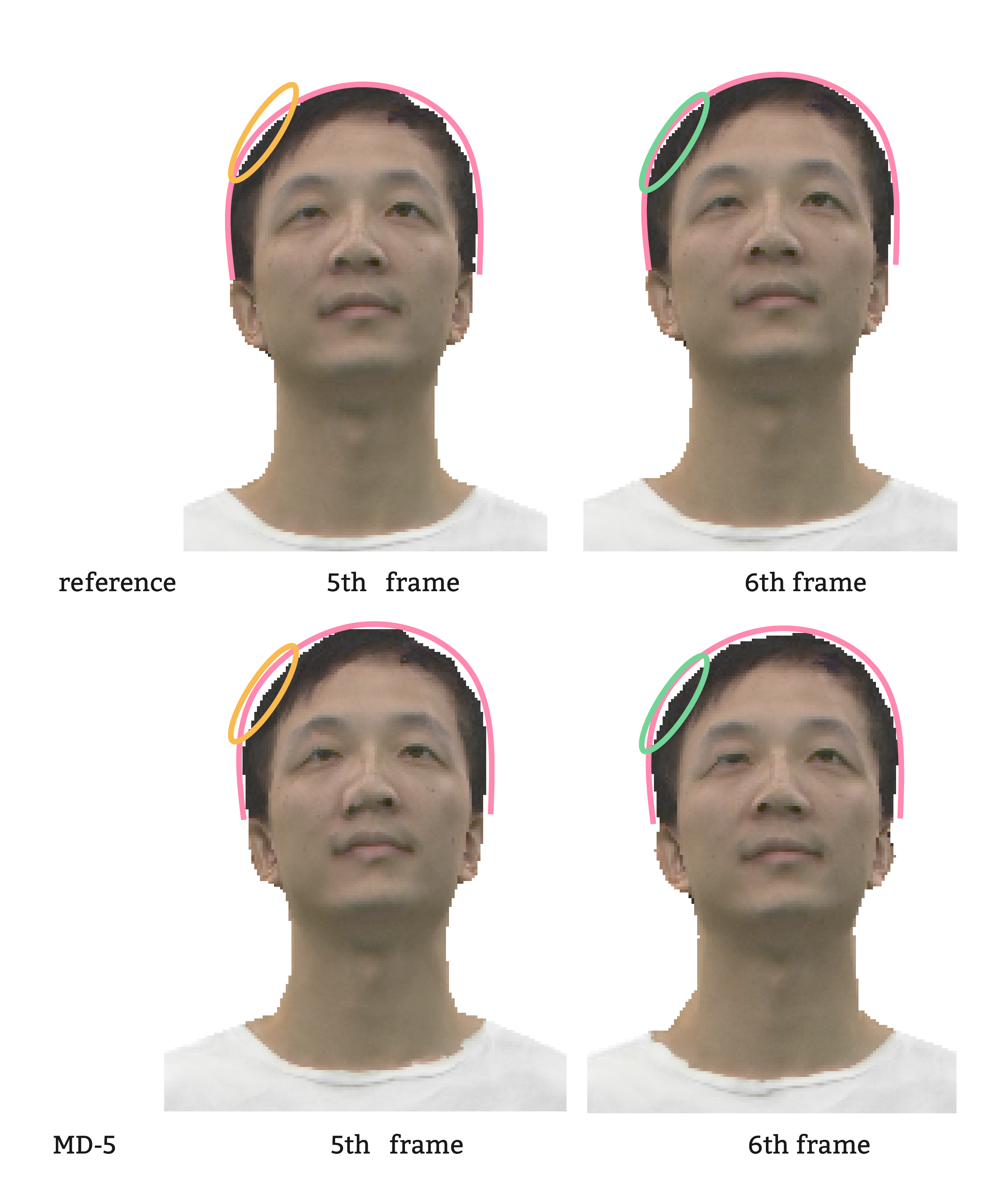}
    \caption{Zoom-in snapshots of MD.}
    \label{fig:dis-md}
\end{figure}

We plot the zoom-in snapshots of the 5th and 6th frames of the reference and the MD-5 distorted ``Basketball\_player'' mesh in Fig. \ref{fig:dis-md}.  { First, we can observe that the reference and the MD-5 samples show close subjective perceptions. This illustrates  5K faces can still exhibit sample characteristics with high fidelity. } Secondly, a pink curve is plotted to mark the general contour of the head for the four examples. We use orange and green ellipses to circle areas having obvious differences between the reference and the MD-5 samples. We find that for adjacent frames, the transition of the reference contours is smoother than that of the MD-5. For the rendered PVSs, MD can lead to contour flicker, and the higher the degree of MD, the more obvious the flicker.

\subsection*{Analysis of TC on Human Perception} 

We hypothesize that the main reason that TC has a limited impact on human perception is that for the subjective experiment setting in this paper, the noticeable distortions caused by TC are relatively small. Moreover, the perceptual distortion is masked due to the frame switching of dynamic sequence. 

\begin{figure}[h]
\setlength{\abovecaptionskip}{0.cm}
\setlength{\belowcaptionskip}{-0.cm}
    \centering
    \includegraphics[width=1\linewidth]{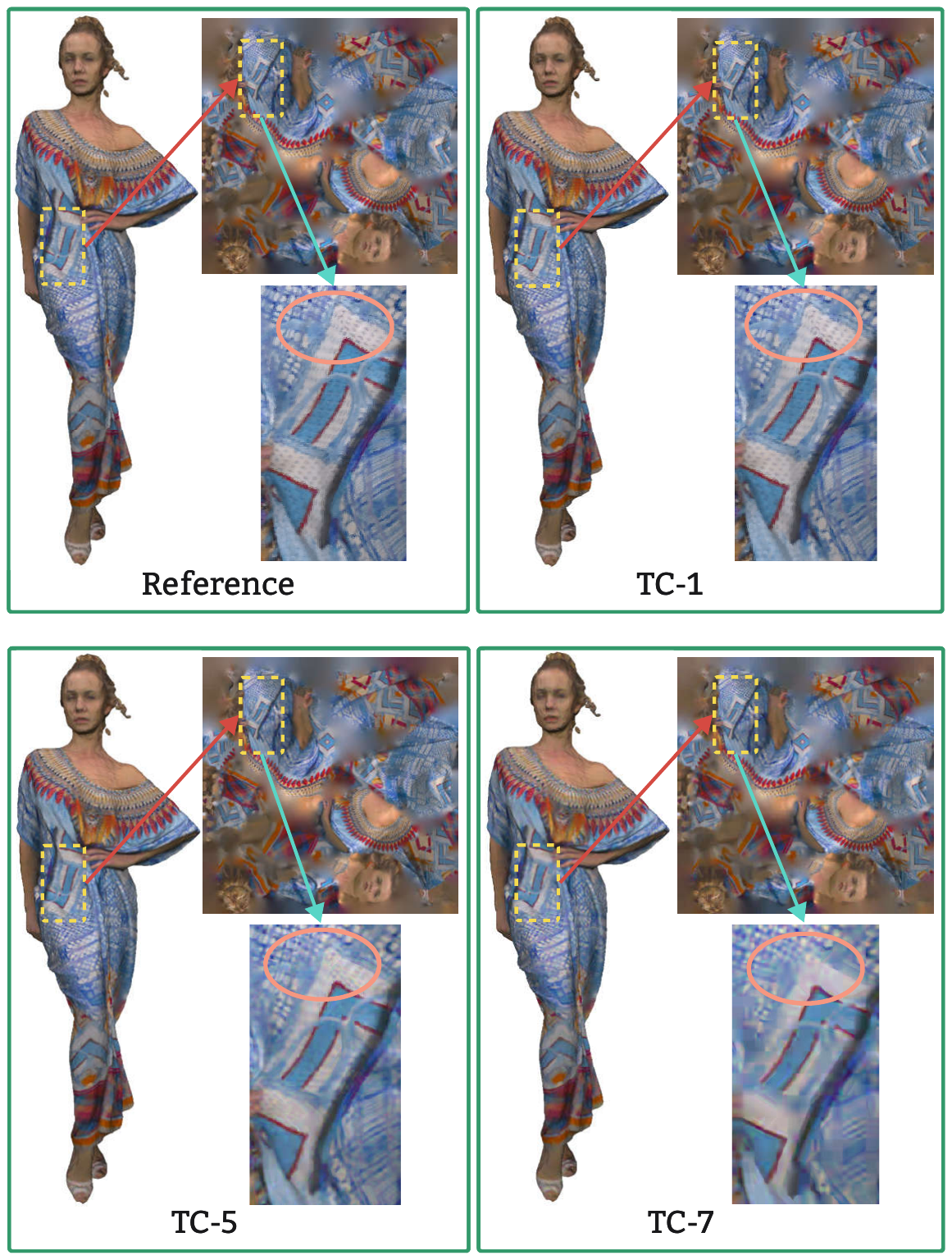}
    \caption{Snapshots of distortion TC.}
    \label{fig:dis-tc}
\end{figure}

We illustrate the global and zoom-in snapshots of the ``Longdress'' reference, TC-1, TC-5 and TC-7 in Fig. \ref{fig:dis-tc}. For the global snapshots, we cannot perceive obvious differences. For the zoom-in snapshots of the texture map, however, we can see that TC causes a slight blurring of the  texture (cf., the content in the orange ellipses), which is limited even for the distortion caused by the maximum QP value (TC-7: QP = 50). Besides, the frame switching of dynamic sequences tends to mask the visibility of fine textures, which makes it hard for the viewers to distinguish the quality degradation, making the TC distortions  more obvious for static colored meshes. Therefore, the viewers usually give a high score for TC samples.
%




\ifCLASSOPTIONcaptionsoff
  \newpage
\fi



\bibliographystyle{IEEEtran}
\bibliography{ref}

\begin{thebibliography}{10}
\providecommand{\url}[1]{#1}
\csname url@samestyle\endcsname
\providecommand{\newblock}{\relax}
\providecommand{\bibinfo}[2]{#2}
\providecommand{\BIBentrySTDinterwordspacing}{\spaceskip=0pt\relax}
\providecommand{\BIBentryALTinterwordstretchfactor}{4}
\providecommand{\BIBentryALTinterwordspacing}{\spaceskip=\fontdimen2\font plus
\BIBentryALTinterwordstretchfactor\fontdimen3\font minus
  \fontdimen4\font\relax}
\providecommand{\BIBforeignlanguage}[2]{{%
\expandafter\ifx\csname l@#1\endcsname\relax
\typeout{** WARNING: IEEEtran.bst: No hyphenation pattern has been}%
\typeout{** loaded for the language `#1'. Using the pattern for}%
\typeout{** the default language instead.}%
\else
\language=\csname l@#1\endcsname
\fi
#2}}
\providecommand{\BIBdecl}{\relax}
\BIBdecl

\bibitem{attene2013polygon}
M.~Attene, M.~Campen, and L.~Kobbelt, ``Polygon mesh repairing: An application
  perspective,'' \emph{ACM Computing Surveys}, vol.~45, no.~2, pp. 1--33, 2013.

\bibitem{abouelaziz2017mesh}
I.~Abouelaziz, A.~Chetouani, M.~El~Hassouni, and H.~Cherifi, ``Mesh visual
  quality assessment metrics: A comparison study,'' \emph{IEEE Int. Conf.
  Signal-Image Technology \& Internet-Based Systems}, pp. 283--288, 2017.

\bibitem{MPEG-MESH-cfp}
WG7, ``Cfp for dynamic mesh coding,'' \emph{ISO/IEC JTC 1/SC 29/WG 7, MPEG 3D
  Graphics Coding, WG7 N0231}, 2021.

\bibitem{meshcomp-maglo20153d}
A.~Maglo, G.~Lavoué, F.~Dupont, and C.~Hudelot, ``3d mesh compression: survey,
  comparisons, and emerging trends,'' \emph{ACM Computing Surveys}, vol.~47,
  no.~3, pp. 1--41, 2015.

\bibitem{mesht-yang2004progressive}
S.~Yang, C.-S. Kim, and C.-C. Kuo, ``A progressive view-dependent technique for
  interactive 3-d mesh transmission,'' \emph{IEEE Trans. Circuits and Systems
  for Video Technology}, vol.~14, no.~11, pp. 1249--1264, 2004.

\bibitem{meshre-jiang20183d}
L.~Jiang, J.~Zhang, B.~Deng, H.~Li, and L.~Liu, ``3d face reconstruction with
  geometry details from a single image,'' \emph{IEEE Trans. Image Processing},
  vol.~27, no.~10, pp. 4756--4770, 2018.

\bibitem{meshen-hansen2005mesh}
G.~A. Hansen, R.~W. Douglass, and A.~Zardecki, ``Mesh enhancement: selected
  elliptic methods, foundations and applications,'' \emph{World Scientific},
  2005.

\bibitem{meshcompression-christaki2019subjective}
K.~Christaki, E.~Christakis, P.~Drakoulis, A.~Doumanoglou, N.~Zioulis,
  D.~Zarpalas, and P.~Daras, ``Subjective visual quality assessment of
  immersive 3d media compressed by open-source static 3d mesh codecs,''
  \emph{Int. Conf. MultiMedia Modeling}, pp. 80--91, 2019.

\bibitem{DWPM-corsini2007watermarked}
M.~Corsini, E.~D. Gelasca, T.~Ebrahimi, and M.~Barni, ``Watermarked 3-d mesh
  quality assessment,'' \emph{IEEE Trans. Multimedia}, vol.~9, no.~2, pp.
  247--256, 2007.

\bibitem{guo2016subjective}
J.~Guo, V.~Vidal, I.~Cheng, A.~Basu, A.~Baskurt, and G.~Lavoue, ``Subjective
  and objective visual quality assessment of textured 3d meshes,'' \emph{ACM
  Trans. Applied Perception}, vol.~14, no.~2, pp. 1--20, 2016.

\bibitem{dy-torkhani2015perceptual}
F.~Torkhani, K.~Wang, and J.-M. Chassery, ``Perceptual quality assessment of 3d
  dynamic meshes: Subjective and objective studies,'' \emph{Signal Processing:
  Image Communication}, vol.~31, pp. 185--204, 2015.

\bibitem{TMM-masking}
A.~Javaheri, C.~Brites, F.~Pereira, and J.~Ascenso, ``Point cloud rendering
  after coding: Impacts on subjective and objective quality,'' \emph{IEEE
  Trans. Multimedia}, vol.~23, pp. 4049--4064, 2021.

\bibitem{MSDM-lavoue2006perceptually}
G.~Lavoué, E.~D. Gelasca, F.~Dupont, A.~Baskurt, and T.~Ebrahimi,
  ``Perceptually driven 3d distance metrics with application to watermarking,''
  \emph{Applications of Digital Image Processing XXIX}, vol. 6312, pp.
  150--161, 2006.

\bibitem{dame-vavsa2012dihedral}
L.~Vasa and J.~Rus, ``Dihedral angle mesh error: a fast perception correlated
  distortion measure for fixed connectivity triangle meshes,'' \emph{Computer
  Graphics Forum}, vol.~31, no.~5, pp. 1715--1724, 2012.

\bibitem{MPEG-MESH-Coding}
WG7, ``Anchors for dynamic mesh coding evaluation,'' \emph{ISO/IEC JTC 1/SC
  29/WG 7, MPEG 3D Graphics Coding, WG7 N0278}, 2022.

\bibitem{MPEG-MESH-metric}
------, ``Metrics for dynamic mesh coding,'' \emph{ISO/IEC JTC 1/SC 29/WG 7,
  MPEG 3D Graphics Coding, WG7 N0225}, 2021.

\bibitem{wang2004image}
Z.~Wang, A.~C. Bovik, H.~R. Sheikh, and E.~P. Simoncelli, ``Image quality
  assessment: from error visibility to structural similarity,'' \emph{IEEE
  Trans. Image Processing}, vol.~13, no.~4, pp. 600--612, 2004.

\bibitem{vmaf-li2016toward}
Z.~Li, A.~Aaron, I.~Katsavounidis, A.~Moorthy, and M.~Manohara, ``Toward a
  practical perceptual video quality metric,'' \emph{The Netflix Tech Blog},
  vol.~6, no.~2, p.~2, 2016.

\bibitem{nehme2020visual}
Y.~Nehmé, F.~Dupont, J.-P. Farrugia, P.~Le~Callet, and G.~Lavoué, ``Visual
  quality of 3d meshes with diffuse colors in virtual reality: Subjective and
  objective evaluation,'' \emph{IEEE Trans. Visualization and Computer
  Graphics}, vol.~27, no.~3, pp. 2202--2219, 2020.

\bibitem{nehme2022textured}
Y.~Nehmé, J.~Delanoy, F.~Dupont, J.-P. Farrugia, P.~Le~Callet, and G.~Lavoué,
  ``Textured mesh quality assessment: Large-scale dataset and deep
  learning-based quality metric,'' \emph{ACM Trans. Graphics}, vol.~42, no.~3,
  pp. 1--20, 2023.

\bibitem{lavoue2006perceptually}
G.~Lavoué, E.~D. Gelasca, F.~Dupont, A.~Baskurt, and T.~Ebrahimi,
  ``Perceptually driven 3d distance metrics with application to watermarking,''
  \emph{Applications of Digital Image Processing XXIX}, vol. 6312, pp.
  150--161, 2006.

\bibitem{rms-cignoni1998metro}
P.~Cignoni, C.~Rocchini, and R.~Scopigno, ``Metro: measuring error on
  simplified surfaces,'' \emph{Computer Graphics Forum}, vol.~17, no.~2, pp.
  167--174, 1998.

\bibitem{hd-aspert2002mesh}
N.~Aspert, D.~Santa-Cruz, and T.~Ebrahimi, ``Mesh: Measuring errors between
  surfaces using the hausdorff distance,'' \emph{Proc. Int. Conf. Multimedia
  and Expo}, vol.~1, pp. 705--708, 2002.

\bibitem{GL-karni2000spectral}
Z.~Karni and C.~Gotsman, ``Spectral compression of mesh geometry,'' \emph{Proc.
  conf. Computer Graphics and Interactive Techniques}, pp. 279--286, 2000.

\bibitem{pro-lavoue2015efficiency}
G.~Lavoué, M.~C. Larabi, and L.~Vasa, ``On the efficiency of image metrics for
  evaluating the visual quality of 3d models,'' \emph{IEEE Trans. Visualization
  and Computer Graphics}, vol.~22, no.~8, pp. 1987--1999, 2015.

\bibitem{fsim-zhang2011fsim}
L.~Zhang, L.~Zhang, X.~Mou, and D.~Zhang, ``Fsim: A feature similarity index
  for image quality assessment,'' \emph{IEEE Trans. Image Processing}, vol.~20,
  no.~8, pp. 2378--2386, 2011.

\bibitem{fibonacci}
M.~Roberts, ``How to evenly distribute points on a sphere more effectively than
  the canonical fibonacci lattice,''
  \url{http://extremelearning.com.au/evenly-distributing-points-on-a-sphere/}.

\bibitem{p-nr2-abouelaziz20203d}
I.~Abouelaziz, A.~Chetouani, M.~El~Hassouni, L.~J. Latecki, and H.~Cherifi,
  ``3d visual saliency and convolutional neural network for blind mesh quality
  assessment,'' \emph{Neural Computing and Applications}, vol.~32, pp.
  16\,589--16\,603, 2020.

\bibitem{p-n4-abouelaziz2020combination}
------, ``Combination of handcrafted and deep learning-based features for 3d
  mesh quality assessment,'' \emph{IEEE Int. Conf. Image Processing}, pp.
  171--175, 2020.

\bibitem{p-n5-abouelaziz2018convolutional}
------, ``Convolutional neural network for blind mesh visual quality assessment
  using 3d visual saliency,'' \emph{IEEE Int. Conf. Image Processing}, pp.
  3533--3537, 2018.

\bibitem{yang2020predicting}
Q.~Yang, H.~Chen, Z.~Ma, Y.~Xu, R.~Tang, and J.~Sun, ``Predicting the
  perceptual quality of point cloud: A 3d-to-2d projection-based exploration,''
  \emph{IEEE Trans. Multimedia}, vol.~23, pp. 3877--3891, 2020.

\bibitem{meynet2020pcqm}
G.~Meynet, Y.~Nehmé, J.~Digne, and G.~Lavoué, ``Pcqm: A full-reference
  quality metric for colored 3d point clouds,'' in \emph{IEEE Int. Conf.
  Quality of Multimedia Experience}, 2020, pp. 1--6.

\bibitem{yang2020inferring}
Q.~Yang, Z.~Ma, Y.~Xu, Z.~Li, and J.~Sun, ``Inferring point cloud quality via
  graph similarity,'' \emph{IEEE Trans. Pattern Analysis and Machine
  Intelligence}, vol.~44, no.~6, pp. 3015--3029, 2022.

\bibitem{Yang_2022_CVPR}
Q.~Yang, Y.~Liu, S.~Chen, Y.~Xu, and J.~Sun, ``No-reference point cloud quality
  assessment via domain adaptation,'' \emph{Proc. IEEE/CVF Conf. Computer
  Vision and Pattern Recognition}, pp. 21\,179--21\,188, June 2022.

\bibitem{yang2021mped}
Q.~Yang, Y.~Zhang, S.~Chen, Y.~Xu, J.~Sun, and Z.~Ma, ``Mped: Quantifying point
  cloud distortion based on multiscale potential energy discrepancy,''
  \emph{IEEE Trans. Pattern Analysis and Machine Intelligence}, pp. 1--18,
  2022.

\bibitem{shanTVCG}
Z.~Shan, Q.~Yang, R.~Ye, Y.~Zhang, Y.~Xu, X.~Xu, and S.~Liu,
  ``Gpa-net:no-reference point cloud quality assessment with multi-task graph
  convolutional network,'' \emph{IEEE Trans. Visualization and Computer
  Graphics}, pp. 1--13, 2023.

\bibitem{suTIP}
H.~Su, Q.~Liu, Y.~Liu, H.~Yuan, H.~Yang, Z.~Pan, and Z.~Wang, ``Bitstream-based
  perceptual quality assessment of compressed 3d point clouds,'' \emph{IEEE
  Trans. Image Processing}, vol.~32, pp. 1815--1828, 2023.

\bibitem{liuTIP}
Q.~Liu, H.~Yuan, R.~Hamzaoui, H.~Su, J.~Hou, and H.~Yang, ``Reduced reference
  perceptual quality model with application to rate control for video-based
  point cloud compression,'' \emph{IEEE Trans. Image Processing}, vol.~30, pp.
  6623--6636, 2021.

\bibitem{liuTVCG}
Q.~Liu, H.~Su, Z.~Duanmu, W.~Liu, and Z.~Wang, ``Perceptual quality assessment
  of colored 3d point clouds,'' \emph{IEEE Trans. Visualization and Computer
  Graphics}, vol.~29, no.~8, pp. 3642--3655, 2023.

\bibitem{8i}
E.~d’Eon, B.~Harrison, T.~Myers, and P.~A. Chou, ``8i voxelized full bodies
  – a voxelized point cloud dataset,'' \emph{ISO/IEC JTC1/SC29 Joint WG11/WG1
  (MPEG/JPEG) input document M40059/M74006}, 2017.

\bibitem{8i-m}
D.~Graziosi, A.~Zaghetto, and A.~Tabatabai, ``V-pcc][ee2.6-related] mesh
  generation script update,'' \emph{ISO/IEC JTC1/SC29/WG7 M55366}, 2017.

\bibitem{owlii}
Y.~Xu, Y.~Lu, and Z.~Wen, ``Owlii dynamic human mesh sequence dataset,''
  \emph{ISO/IEC JTC1/SC29/WG11 M41658}, 2017.

\bibitem{xdprod}
R.~Schaefer, P.~Andrivon, J.~Ricard, and C.~Guede, ``Volucap and xd productions
  datasets,'' \emph{ISO/IEC JTC1/SC29/WG7 M56192}, 2021.

\bibitem{vsense}
R.~Pagés, E.~Zerman, K.~Amplianitis, J.~Ondrej, and A.~Smolic, ``Volograms \&
  vsense volumetric video dataset,'' \emph{ISO/IEC JTC1/SC29/WG7 M56767}, 2021.

\bibitem{draco}
Google, ``Draco library for 3d geometric meshes and point cloud compression,''
  \emph{\url{https://google.github.io/draco/}}.

\bibitem{HDRtool}
B.~Williams, ``{HDRTools software},''
  \emph{\url{https://gitlab.com/standards/HDRTools.git}}.

\bibitem{HEVC}
K.~Suehring, ``{HEVC reference software},''
  \emph{\url{https://vcgit.hhi.fraunhofer.de/jvet/HM/-/tags/HM-16.21+SCM-8.8}}.

\bibitem{ffmpeg}
FFMPEG, ``{A complete, cross-platform solution to record, convert and stream
  audio and video.}'' \emph{\url{https://ffmpeg.org/}}.

\bibitem{BT500}
B.~ITU-R~RECOMMENDATION, ``Methodology for the subjective assessment of the
  quality of television pictures,'' \emph{International Telecommunication
  Union}, 2002.

\bibitem{11scale-rating}
P.~ITU-T~RECOMMENDATION, ``Subjective video quality assessment methods for
  multimedia applications,'' \emph{International Telecommunication Union},
  1999.

\bibitem{PintVR}
ITU-T, ``Subjective test method for interactive virtual reality applications,''
  \emph{\url{https://www.itu.int/ITU-T/workprog/wp_item.aspx?isn=17817}}.

\bibitem{open3d}
Q.-Y. Zhou, J.~Park, and V.~Koltun, ``{Open3D}: {A} modern library for {3D}
  data processing,'' \emph{arXiv:1801.09847}, 2018.

\bibitem{m57896}
J.~Jung, M.~Wien, and V.~Baronini, ``Draft guidelines for remote experts
  viewing sessions (v2),'' \emph{ISO/IEC JTC 1/SC 29/AG 5 M57896}, 2021.

\bibitem{antsiferova2022video}
A.~Antsiferova, S.~Lavrushkin, M.~Smirnov, A.~Gushchin, D.~S. Vatolin, and
  D.~Kulikov, ``Video compression dataset and benchmark of learning-based
  video-quality metrics,'' \emph{Conf. Neural Information Processing Systems
  Datasets and Benchmarks Track}, 2022.

\bibitem{wang2003multiscale}
Z.~Wang, E.~P. Simoncelli, and A.~C. Bovik, ``Multiscale structural similarity
  for image quality assessment,'' \emph{IEEE Asilomar Conf. Signals, Systems \&
  Computers}, vol.~2, pp. 1398--1402, 2003.

\bibitem{vqm}
M.~Pinson and S.~Wolf, ``A new standardized method for objectively measuring
  video quality,'' \emph{IEEE Trans. Broadcasting}, vol.~50, no.~3, pp.
  312--322, 2004.

\bibitem{li2009three}
C.~Li and A.~C. Bovik, ``Three-component weighted structural similarity
  index,'' \emph{Image quality and system performance VI}, vol. 7242, pp.
  252--260, 2009.

\bibitem{video2003final}
VQEG, ``Final report from the video quality experts group on the validation of
  objective models of video quality assessment,''
  \emph{http://www.its.bldrdoc.gov/vqeg/vqeg-home.aspx}.

\bibitem{yang2020msea}
Q.~Yang, Z.~Ma, Y.~Xu, L.~Yang, W.~Zhang, and J.~Sun, ``Modeling the screen
  content image quality via multiscale edge attention similarity,'' \emph{IEEE
  Trans. Broadcasting}, vol.~66, no.~2, pp. 310--321, 2020.

\bibitem{alexiou2019exploiting}
E.~Alexiou and T.~Ebrahimi, ``Exploiting user interactivity in quality
  assessment of point cloud imaging,'' \emph{IEEE Int. Conf. Quality of
  Multimedia Experience}, pp. 1--6, 2019.

\bibitem{icme2023}
K.~Yang, Q.~Yang, J.~Jung, Y.~Xu, X.~Xu, and S.~Liu, ``Exploring the influence
  of view and camera path selection for dynamic mesh quality assessment,''
  \emph{IEEE Int. Conf. Multimedia and Expo}, 2023.

\bibitem{fu2023surface}
C.~Fu, X.~Zhang, T.~Nguyen-Canh, X.~Xu, G.~Li, and S.~Liu, ``Surface-sampling
  based objective quality assessment metrics for meshes,'' in \emph{IEEE Int.
  Conf. Acoustics, Speech and Signal Processing}, 2023, pp. 1--5.

\end{thebibliography}
\end{document}